\documentclass[journal]{IEEEtran}

\usepackage{amssymb,amsmath,amsfonts}
\usepackage{algorithmic}
\usepackage{algorithm}
\usepackage{array}
\usepackage{bbold}
\usepackage{booktabs}
\usepackage{colortbl}
\usepackage[hidelinks]{hyperref}
\usepackage{graphicx}
\usepackage{makecell}
\usepackage[mathlines, switch]{lineno}
\usepackage{mathtools}
\usepackage{multirow}
\usepackage[sort&compress, numbers]{natbib}
\usepackage{pifont}
\usepackage{siunitx}
\usepackage{stfloats}
\usepackage[caption=false,font=footnotesize]{subfig}
\usepackage{textcomp}
\usepackage{xurl}
\usepackage{verbatim}
\usepackage[dvipsnames]{xcolor}

\hypersetup{
  colorlinks   = true, 
  urlcolor     = CadetBlue, 
  linkcolor    = Cerulean, 
  citecolor   = Maroon 
}

\begin{document}

\title{A Stitch in Time Saves Nine: Preserving Policy Compatibility Under Perception Updates in End-to-End Autonomous Driving}

\author{Yueyuan~Li, Yifei~Xiao, Mingyang~Jiang, Xiang~Zuo, Songan~Zhang, and~Ming~Yang 
\thanks{This work is supported in part by the National Natural Science Foundation of China under Grant 62573289, 52402504, and U22A20100 \textit{(Corresponding author: Ming Yang, email: MingYANG@sjtu.edu.cn)}.}
\thanks{Yueyuan Li, Yifei Xiao, Mingyang Jiang, and Ming Yang are with the School of Automation and Intelligent Sensing, Shanghai Jiao Tong University, Key Laboratory of System Control and Information Processing, Ministry of Education of China, Shanghai, 200240, CN.}
\thanks{Xiang Zuo is with the School of Aeronautics and Astronautics, Shanghai Jiao Tong University, Shanghai, 200240, CN.}
\thanks{Songan Zhang is with the Global Institute of Future Technology, Shanghai Jiao Tong University, Shanghai, 200240, CN.}
}

\maketitle


\begin{abstract}
End-to-end autonomous driving systems tightly couple perception and decision-making through latent representations.
Consequently, updates to perception models can alter these representations and degrade the performance of downstream policies that remain fixed.
Existing solutions typically rely on policy retraining or architectural decoupling, both of which incur substantial computation and validation costs.
In this paper, we formulate the model stitching problem for end-to-end autonomous driving and test the hypothesis that policy compatibility can be preserved through lightweight latent-space alignment.
We study low-complexity model stitching methods, including linear and convolutional stitchers, for restoring compatibility between updated perception modules and frozen downstream policy modules.
Experiments demonstrate that stitching effectively preserves downstream driving behavior under diverse perception updates, including changes in random initialization, sensor configuration, and training domain.
In the most challenging cross-domain setting from nuScenes to CARLA, convolutional stitching retains over 91\% of the no-shift driving score while reducing adaptation time from \SI{22.18}{h} to \SI{0.91}{h}.
These results suggest that model stitching provides an effective and computationally efficient alternative to retraining or fine-tuning for maintaining end-to-end autonomous driving systems.
The model will be open-sourced upon paper acceptance at \url{https://github.com/SCP-CN-001/model-stitching} to support further research and development in autonomous driving.
\end{abstract}



\section{Introduction}

Modern transportation systems are undergoing continuous and rapid evolution.
Global electric car sales exceeded 20 million in 2025, while the number of available electric car models more than doubled over the past five years and reached nearly 1,000 in 2025~\cite{IEA2026GlobalEVOutlook}.
As vehicles with diverse sensing suites, software stacks, and automation capabilities coexist on the road, autonomous driving systems (ADSs) must operate under heterogeneous and evolving deployment conditions.

As a key component of intelligent transportation systems, ADSs are required to perceive dynamic traffic environments, interact with surrounding road users, and generate safe driving decisions.
Recent end-to-end ADS frameworks learn driving policies directly from high-dimensional sensor observations by jointly optimizing perception and decision-making modules~\cite{tampuu2020survey}.
Compared with conventional modular pipelines, this paradigm reduces handcrafted interfaces, mitigates error accumulation, and improves computational efficiency~\cite{chen2024end}.

However, the tight coupling that enables end-to-end optimization also makes downstream policies highly sensitive to changes in upstream perception representations~\cite{muller2018driving,chen2024end}.
In practice, perception modules may evolve more frequently due to retraining, sensor modifications, or dataset expansion~\cite{kumar2022fine, liu2023deep}, whereas updating downstream policies is often more difficult because of sparse feedback and safety-critical deployment requirements~\cite{liu2024curse}.
Such asynchronous evolution can shift latent representations outside the distribution expected by fixed downstream policies, challenging the stable maintenance and long-term evolution of end-to-end ADS~\cite{muller2018driving,liang2022effective}.

To address such representation shifts, a common solution is retraining or fine-tuning the downstream policy~\cite{houlsby2019parameter,pfeiffer2020adapterhub,liu2021broad}, but this introduces substantial computation, validation, and deployment costs.
Alternative approaches aim to reduce dependence between perception and planning via structured intermediate representations~\cite{jiang2023vad,lu2025knowledge}, adapters~\cite{jia2023driveadapter}, or decoupled training~\cite{chen2020learning,yang2026worldrft}.
These improve modularity and transferability but usually assume joint optimization or end-to-end validation, which may not hold during routine maintenance with fixed policies and limited data.

\begin{figure}[tb]
    \centering
    \subfloat{\includegraphics[width=0.49\linewidth]{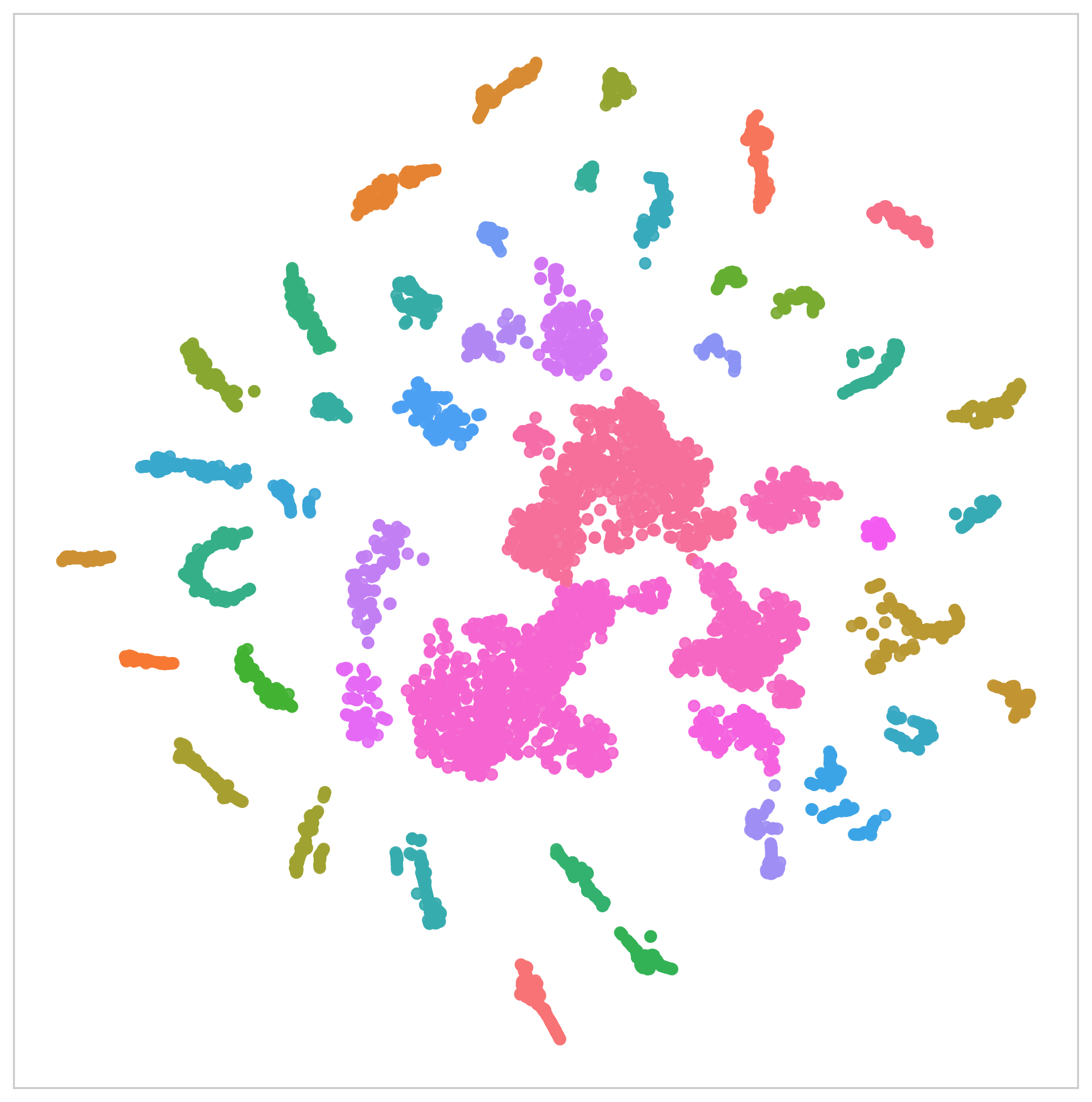}}
    \hfill
    \subfloat{\includegraphics[width=0.49\linewidth]{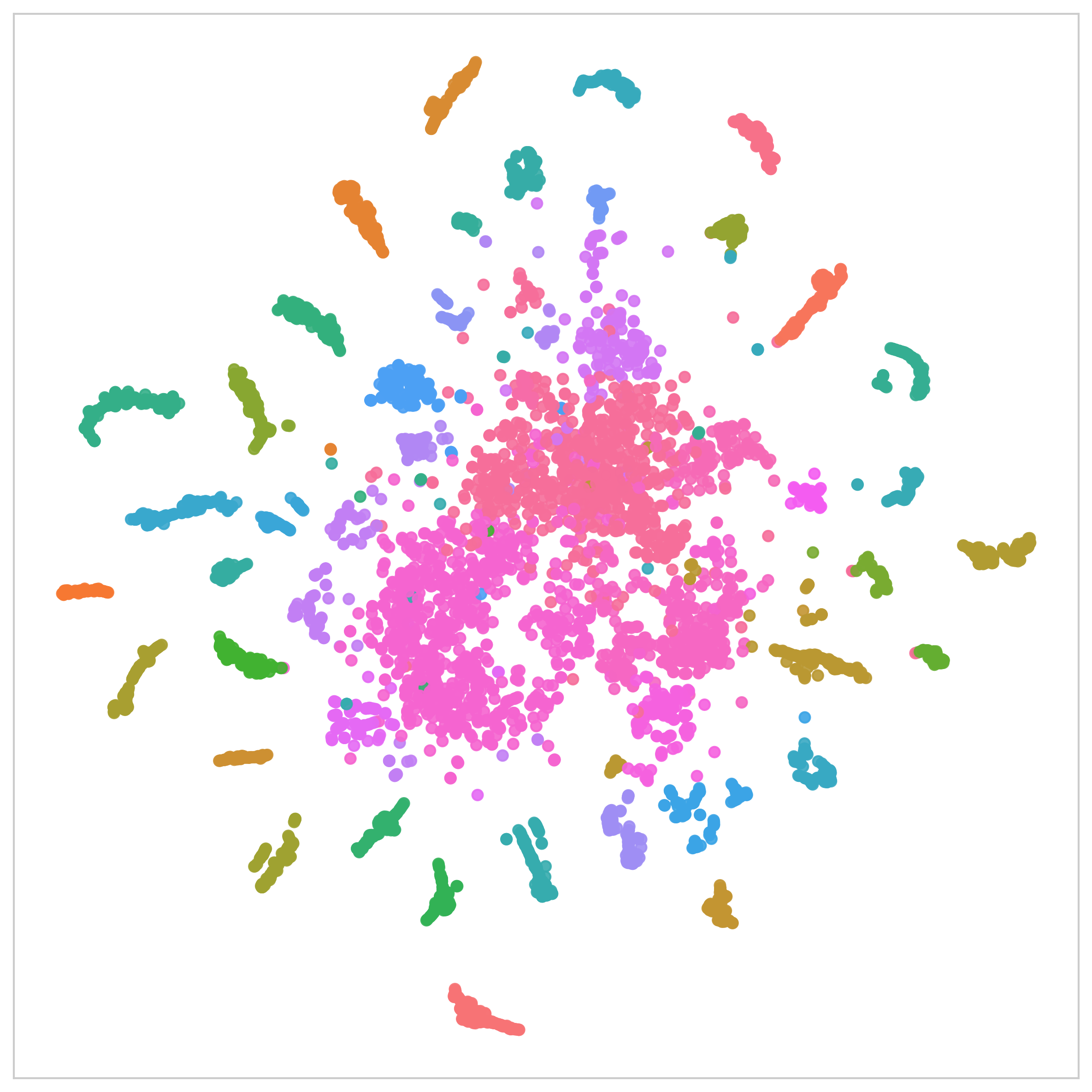}}
    \caption{t-SNE visualizations of latent representations from two perception models trained under the same BEV semantic supervision signal. Colors indicate the same clusters across models.}\label{figure:tSNE-analysis}
\end{figure}

Prior studies show that different neural networks supervised by the same signals can learn related representations~\cite{lenc2015understanding,bansal2021revisiting}, but existing works mainly focus on vision and language.
Whether this compatibility can preserve downstream planning behavior in traffic scenarios remains unclear.
Figure~\ref{figure:tSNE-analysis} illustrates that two perception models trained with the same BEV semantic supervision exhibit similar latent clustering patterns, suggesting that task-relevant structure may be partially preserved across representation shifts.

Rather than retraining the downstream policy, this work formulates the problem as \textbf{repairing the latent interface between a modified perception model and a fixed policy.}
Given an updated perception model, we investigate whether a lightweight stitcher can map the model's latent representation into the space expected by the original downstream model, thereby preserving planning behavior without retraining.

Our main contributions are summarized as follows:
\begin{itemize}

\item We identify and formalize the asynchronous evolution problem between perception and policy in end-to-end autonomous driving, and propose the Policy-Compatible Representation Stitchability Hypothesis.

\item We systematically evaluate lightweight stitching methods under diverse perception shifts, including changes in sensor setup, representation model, training domain.

\item We demonstrate that lightweight latent-space alignment recovers driving behavior more efficiently than retraining and fine-tuning, reducing runtime from \SI{22.18}{h} to \SI{0.91}{h} while updating only 0.001\% of downstream parameters and eliminating the need for online interaction.
\end{itemize}


\section{Related Works}

\subsection{Policy Compatibility under Representation Shifts in End-to-End Driving}

Prior work related to policy compatibility under representation shifts can be viewed through two complementary directions: policy-level adaptation, and training-time design of more structured or transferable perception-policy interfaces.

One direct approach is to further optimize the downstream policy or planner.
Hu et al.~\cite{hu2024pre} combine behavior cloning pre-training with human-guided reinforcement fine-tuning for end-to-end navigation.
Liu et al.~\cite{liu2026reinforced} propose reinforced refinement with self-aware expansion, where specialist policies are refined for hard cases and integrated into a generalist driving model.
Yang et al.~\cite{yang2026worldrft} align latent world-model representations with planning and apply reinforcement fine-tuning to improve safety-critical planning performance.
Although these methods improve policy performance, they assume that the policy or planner can be further optimized, which may not hold in practice due to limited data, safety-critical deployment requirements, or high computational costs.

A second direction addresses perception-policy coupling through training-time interface design.
VAD introduces vectorized scene representations that provide instance-level planning constraints~\cite{jiang2023vad}, while knowledge-driven cognitive representation alignment incorporates explicit decision-relevant structure into the planning representation~\cite{lu2025knowledge}.
Learning by Cheating separates privileged policy learning from sensor-based policy learning through a teacher-student training pipeline~\cite{chen2020learning}.
DriveAdapter further reduces perception-planning coupling by inserting adapters and aligning features between student perception and teacher planning modules~\cite{jia2023driveadapter}.
These methods make the perception-policy interface more structured, transferable, or easier to learn, but their main focus is on improving planning performance, generalization, or training-time decoupling, rather than preserving compatibility with a fixed downstream policy after perception updates.

Together, these two lines of work leave open a deployment-relevant setting: the downstream policy has already been trained and must remain fixed, while the upstream representation changes due to model retraining, dataset updates, or sensor configuration changes.
Our work studies this setting and investigates whether a lightweight stitcher can recover cross-model representation compatibility without modifying the downstream policy.

\subsection{Representation Similarity Analysis and Model Stitching}

Early work by Lenc and Vedaldi examined how different parametrizations of neural networks could capture similar representations, and investigated potential equivalences between these representations~\cite{lenc2015understanding}.
Subsequent studies introduced metrics such as SVCCA, CCA, and CKA to quantify the similarity of representations across different models, layers, and training conditions~\cite{raghu2017svcca,morcos2018insights,kornblith2019similarity}.
Later works also identified important caveats and limitations of these metrics~\cite{balogh2025not,smith2025functional}.
Empirical evidence increasingly shows that representations learned by deep neural networks are often similar across random initializations, training datasets, and even architectures~\cite{csiszarik2021similarity}.
Although representation similarity has been systematically measured, the underlying mechanisms remain largely unexplained.
In the context of this study, these empirical findings indicate that latent spaces can be approximately aligned, motivating the use of a lightweight stitch layer instead of retraining the downstream policy.

More recent research has begun to investigate the causes and potential applications of representation similarity. CoReS~\cite{biondi2023cores} proposed that such similarity arises from a stationary feature space, a hypothesis later corroborated by~\cite{yu2025connecting}.
CoReS also introduced a method for incrementally re-indexing image features as a model is updated.
Nikoorooz et al.~\cite{nikooroo2025cross} demonstrated that structural regularities can induce stable representational geometry even under architectural changes.

In parallel, Bansal et al. formalized the concept of model stitching, where two independently trained models with identical architectures can be connected at a specific layer via a learned linear transformation, and the resulting model can achieve performance comparable to the originals~\cite{bansal2021revisiting}.
This approach directly leverages the similarity of internal representations by aligning feature spaces between models through a linear transformation.
Building on this idea, subsequent works have extended model stitching for transfer learning, cross-architecture alignment, and improved generalization~\cite{maiorca2023latent,teerapittayanon2023stitchnet,anakewat2024self}, demonstrating that representation similarity can be exploited beyond simple observation tasks.
A parallel line of work proposed relative representations~\cite{moschella2022relative}, which align latent spaces through similarity to anchor points and enable zero-shot encoder-decoder stitching, with subsequent extensions improving alignment robustness~\cite{cannistraci2023bootstrapping,cannistraci2023bricks,maiorca2024latent}.

Most existing studies, however, focus on empirical observations in relatively simple vision tasks.
A systematic theoretical understanding of the underlying mechanisms, alongside demonstrations on complex real-world tasks, remains largely absent.
We address this gap by introducing model stitching to end-to-end driving, supported by a formal hypothesis formulation and validation on real sensory data.


\section{Methodology}

\subsection{Problem Formulation}

\begin{figure*}[htb]
    \centering
    \includegraphics[width=\linewidth]{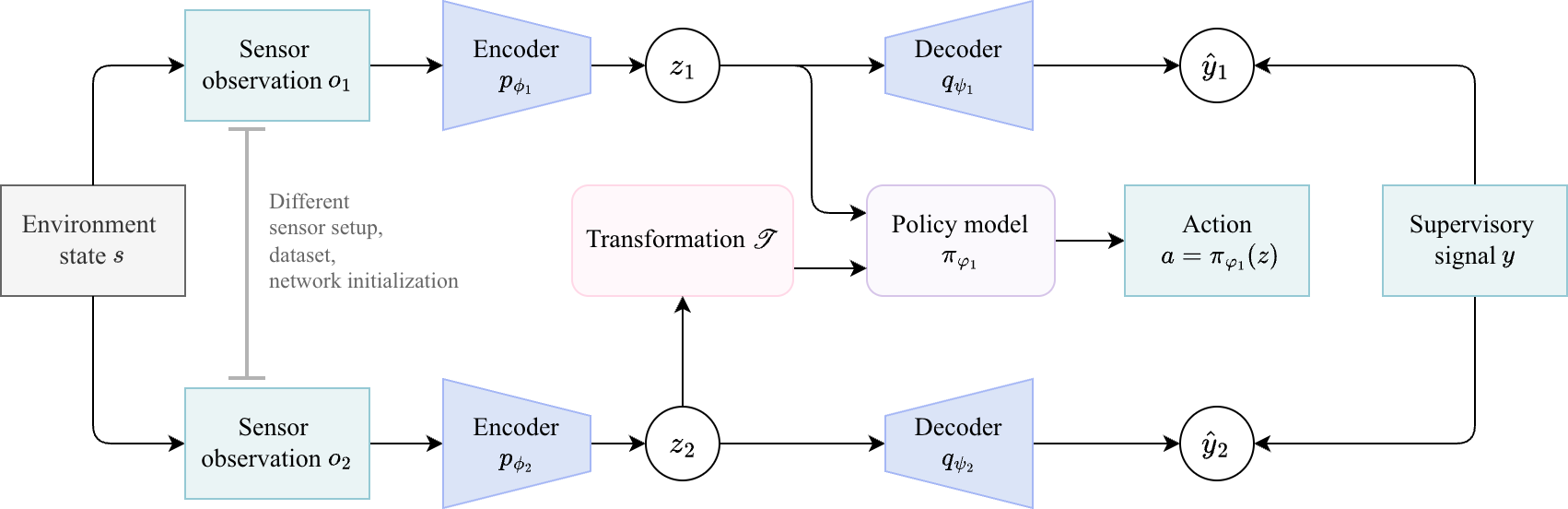}
    \caption{Diagram of the model stitching process.}\label{figure:model-stitching}
\end{figure*}

We formalize model stitching within a typical end-to-end autonomous driving pipeline.
The corresponding workflow is illustrated in Figure~\ref{figure:model-stitching}.
Let $s \in \mathcal{S}$ denote the environment state and $o \in \Omega$ the observation.
A perception encoder model $p_\phi$ maps the observation to a latent representation $z \in \mathcal{Z}$:
\begin{equation}
    z = p_\phi(o), \quad p_\phi\colon \Omega \to \mathcal{Z}.
\end{equation}

In practice, an auxiliary supervisory signal $y \in \mathcal{Y}$, which includes task-relevant information such as state attributes, is often introduced to constrain the learned latent representations~\cite{zeng2019end}:
\begin{equation}
    \hat{y} = q_\psi(z), \quad q_\psi\colon \mathcal{Z} \to \mathcal{Y},
\end{equation}
where $q_\psi$ is a decoder, typically sharing parameters with or jointly optimized with the perception model $p_\phi$.

The decision policy $\pi_\varphi$ takes $z$ and outputs action $a \in \mathcal{A}$:
\begin{equation}
    a = \pi_\varphi(z), \quad \pi_\varphi\colon \mathcal{Z} \to \mathcal{A}.
\end{equation}

Consider two perception models $p_{\phi_1}$ and $p_{\phi_2}$ that differ in parameter initialization, datasets, or sensor configurations:
\begin{equation}
\begin{aligned}
    z_1 &= p_{\phi_1}(o_1), \quad o_1 \in \Omega_1, \quad z_1 \in \mathcal{Z}_1 \subset \mathbb{R}^n \\
    z_2 &= p_{\phi_2}(o_2), \quad o_2 \in \Omega_2, \quad z_2 \in \mathcal{Z}_2 \subset \mathbb{R}^n
\end{aligned}
\end{equation}
For the same environment state $s$, even when trained with supervisory signals of the same type and sharing the same output dimension, the resulting representations may not be directly compatible.

As shown in Figure~\ref{figure:model-stitching}, model stitching aims to find a mapping $\mathcal{T}\colon \mathcal{Z}_2 \to \mathcal{Z}_1$ such that a fixed downstream policy $\pi_{\varphi_1}$ trained on $\mathcal{Z}_1$ can directly consume transformed features from $\mathcal{Z}_2$.
A natural formulation of the stitching objective is to minimize the action-space discrepancy between the original and stitched representations:
\begin{equation}\label{equation:problem-formulation}
\begin{aligned}
    \hat{\mathcal{T}}
    =
    \arg\min_{\mathcal{T}}
    \;
    \mathbb{E}_{(o_1,o_2)\sim\mathcal{D}_{\mathrm{pair}}}
    \Big[
    d_{\mathcal{A}}\big(
    & [\pi_{\varphi_1}\circ p_{\phi_1}](o_1), \\
    & [\pi_{\varphi_1}\circ \mathcal{T}\circ p_{\phi_2}](o_2)
    \big)
    \Big]
\end{aligned}
\end{equation}
where $\pi_{\varphi_1}$ is the downstream policy trained on $\mathcal{Z}_1$,
$\mathcal{D}_{\mathrm{pair}}$ denotes a distribution over paired observations generated from the same underlying environment state $s$, 
and $d_{\mathcal{A}}$ denotes a task-dependent measure.


\subsection{Hypothesis}\label{section:hypothesis}

While end-to-end ADS operates on high-dimensional observations $o$, both the latent representation $z$ and the supervisory signal $y$ aim to capture compact, task-relevant information, such as road topology, traffic participants, and vehicle dynamics.
This suggests that the variations relevant to driving decisions may occupy a much lower-dimensional structure than the raw observation space, as many appearance-level details in $o$ are irrelevant for decision-making.

Consistent with the widely adopted manifold hypothesis in machine learning~\cite{bengio2013representation}, we formalize:

\textbf{Assumption 1 (Task-Relevant Manifold).}
\textit{
Planning-relevant variations of the environment state $s$ reside on a low-dimensional manifold $\mathcal{M}$ within the task-relevant factor space, with intrinsic dimension much smaller than that of the full state space $\mathcal{S}$.
}

This assumption implies that supervision need not account for all information contained in the full state $s$.
What matters is whether the supervisory signal $y$ is informative about the task-relevant factors that parameterize $\mathcal{M}$.
Thus, rather than treating $y$ as a reconstruction target for the environment state, we interpret it as a task-dependent measurement of the aspects of $s$ that should be retained in the learned representation.
In autonomous driving, commonly used intermediate supervision signals, such as object detection, BEV segmentation, and occupancy~\cite{zeng2019end, hu2023planning, hu2022st}, provide such measurements in the form of compact geometric, semantic, and spatial information~\cite{jiang2023vad}.
This motivates the following assumption on the relevance of the supervisory signal to the task.

\textbf{Assumption 2 (Task-Relevant Supervision).}
\textit{
The supervisory signal $y$ is assumed to capture much of the task-relevant information that is consistently useful for downstream planning decisions on the task distribution.
}

We further assume that these supervised factors are retained in the learned latent representation.
This is a decodability assumption: if $z$ preserves the information in $y$, then a lightweight decoder or probe should be able to recover $y$ from $z$ with small error~\cite{lenc2015understanding,bansal2021revisiting,kornblith2019similarity}.

\begin{figure}[tb]
    \centering
    \subfloat[VAEs (initialization)]{\includegraphics[height=2.85cm]{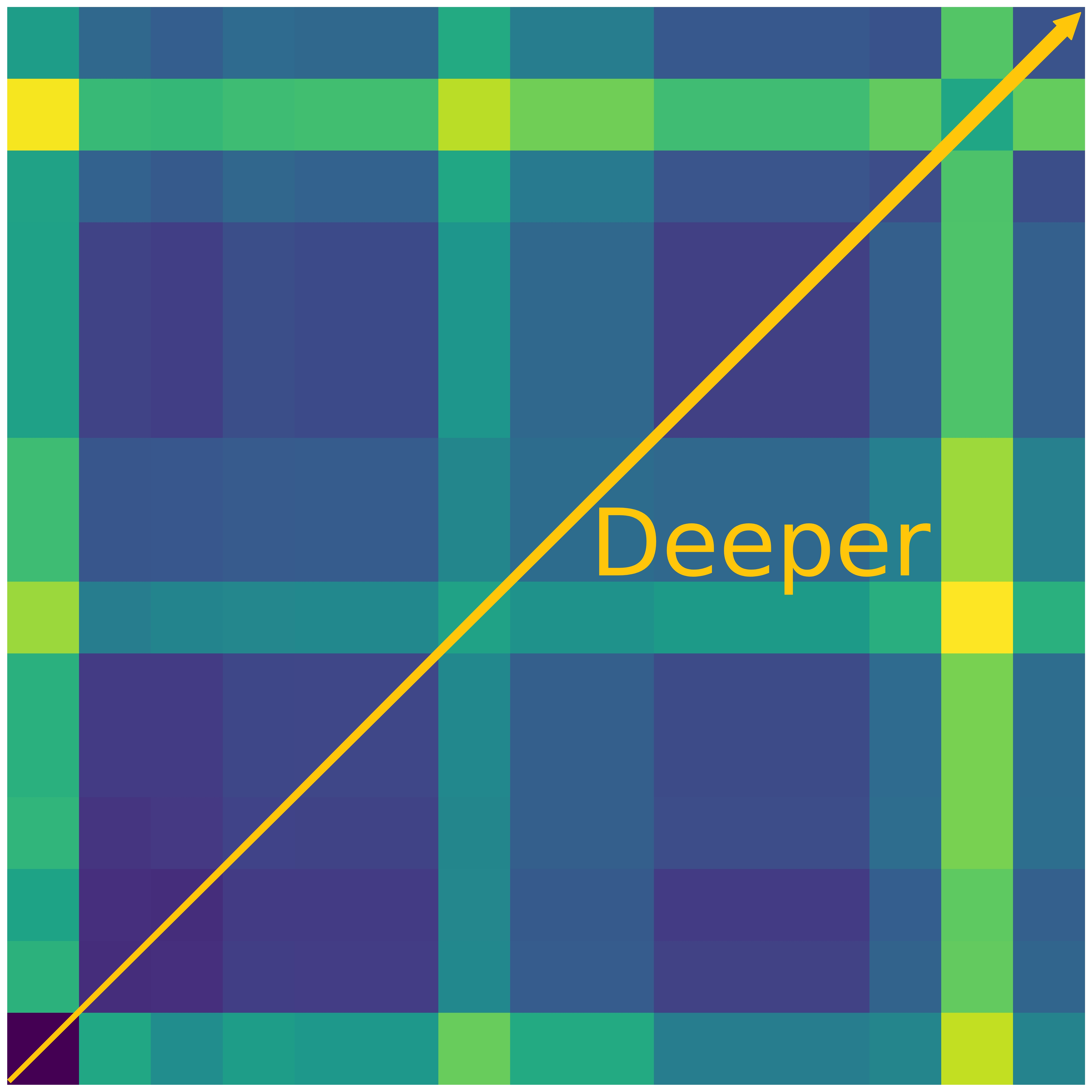}}
    \hfill
    \subfloat[AEs (initialization)]{\includegraphics[height=2.85cm]{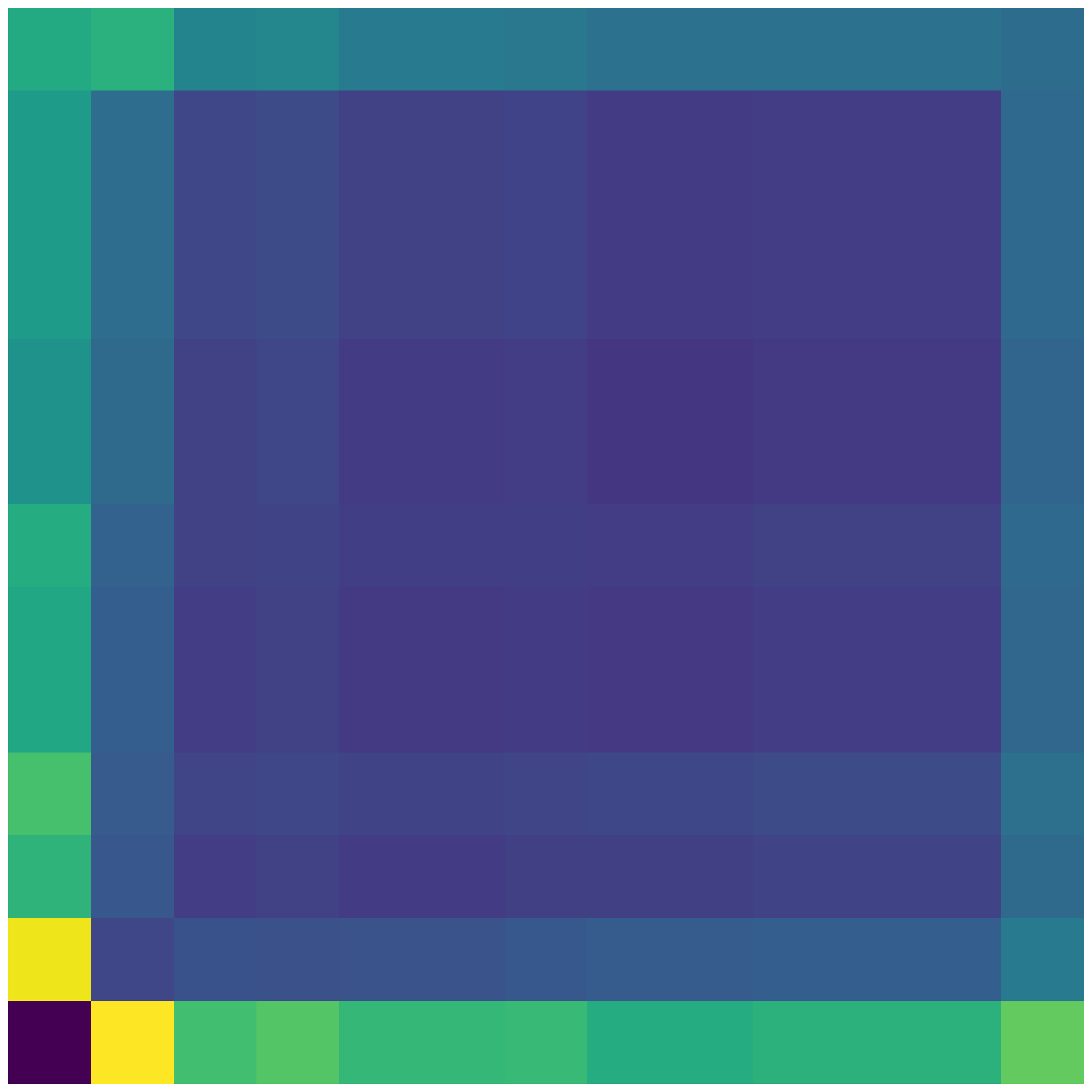}}
    \hfill
    \subfloat[VAE vs. AE (initialization)]{\includegraphics[height=2.85cm]{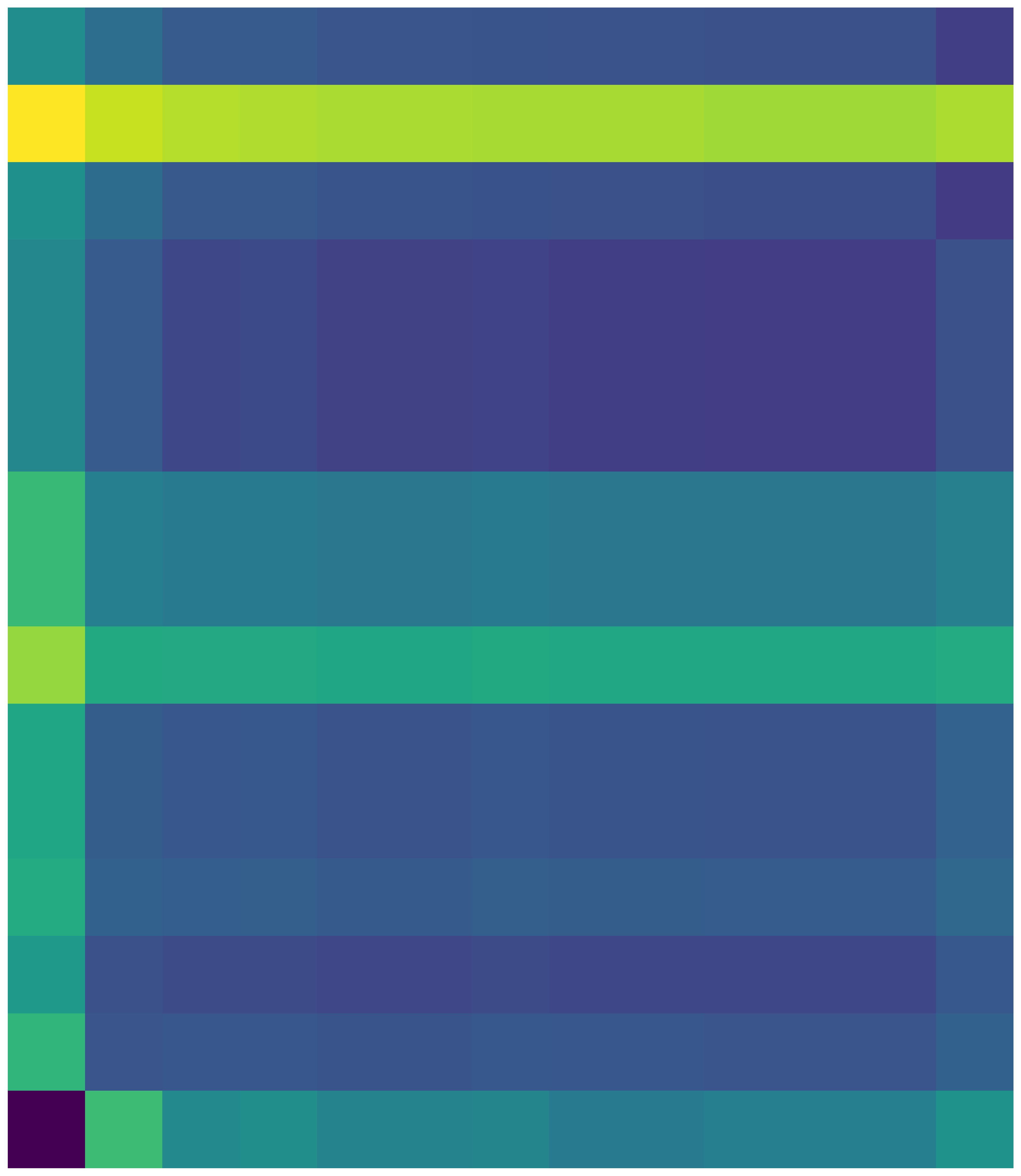}}\\
    \subfloat[VAE vs. VAE (lidar-only)]{\includegraphics[height=2.8cm]{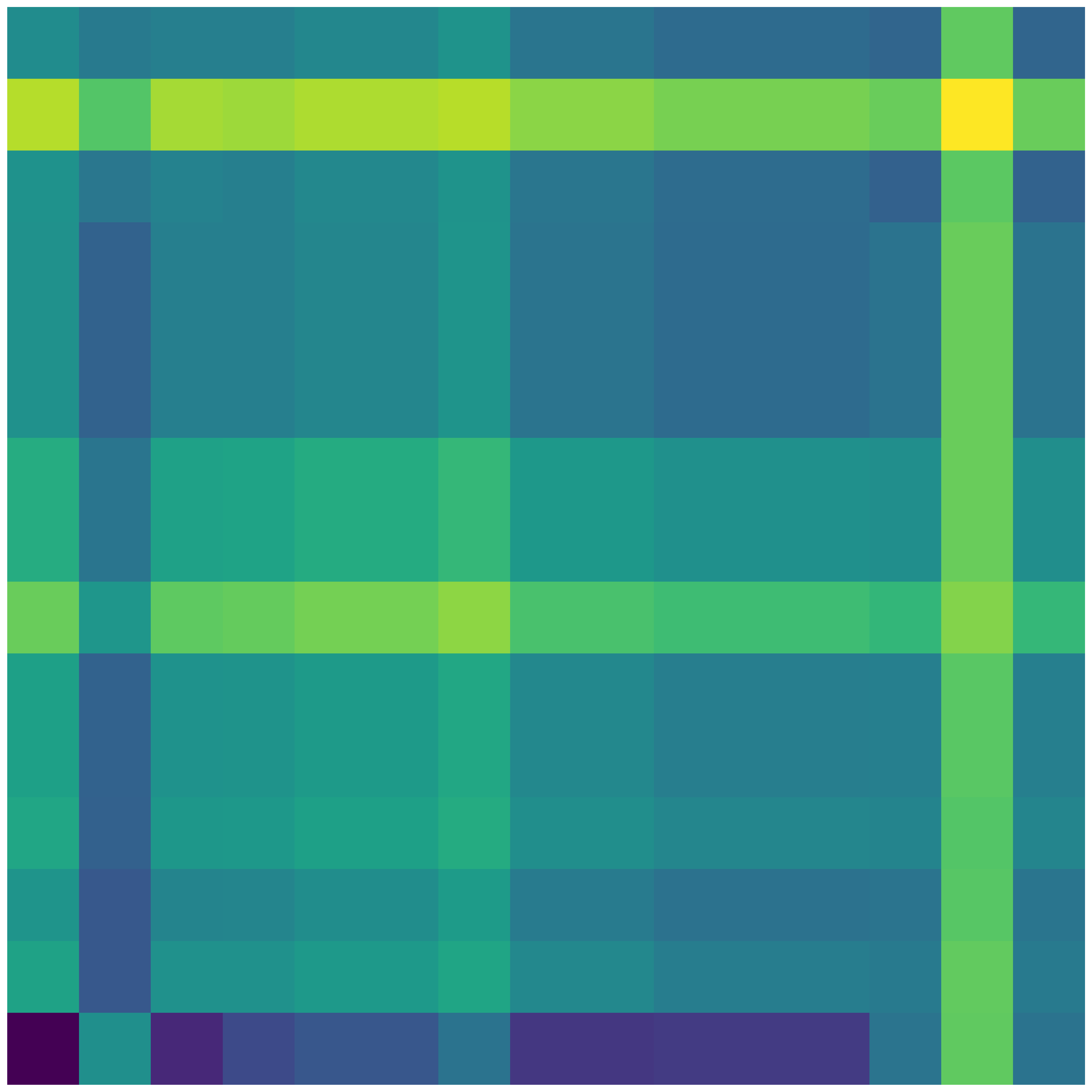}}
    \hfill
    \subfloat[VAE vs. VAE (camera-only)]{\includegraphics[height=2.8cm]{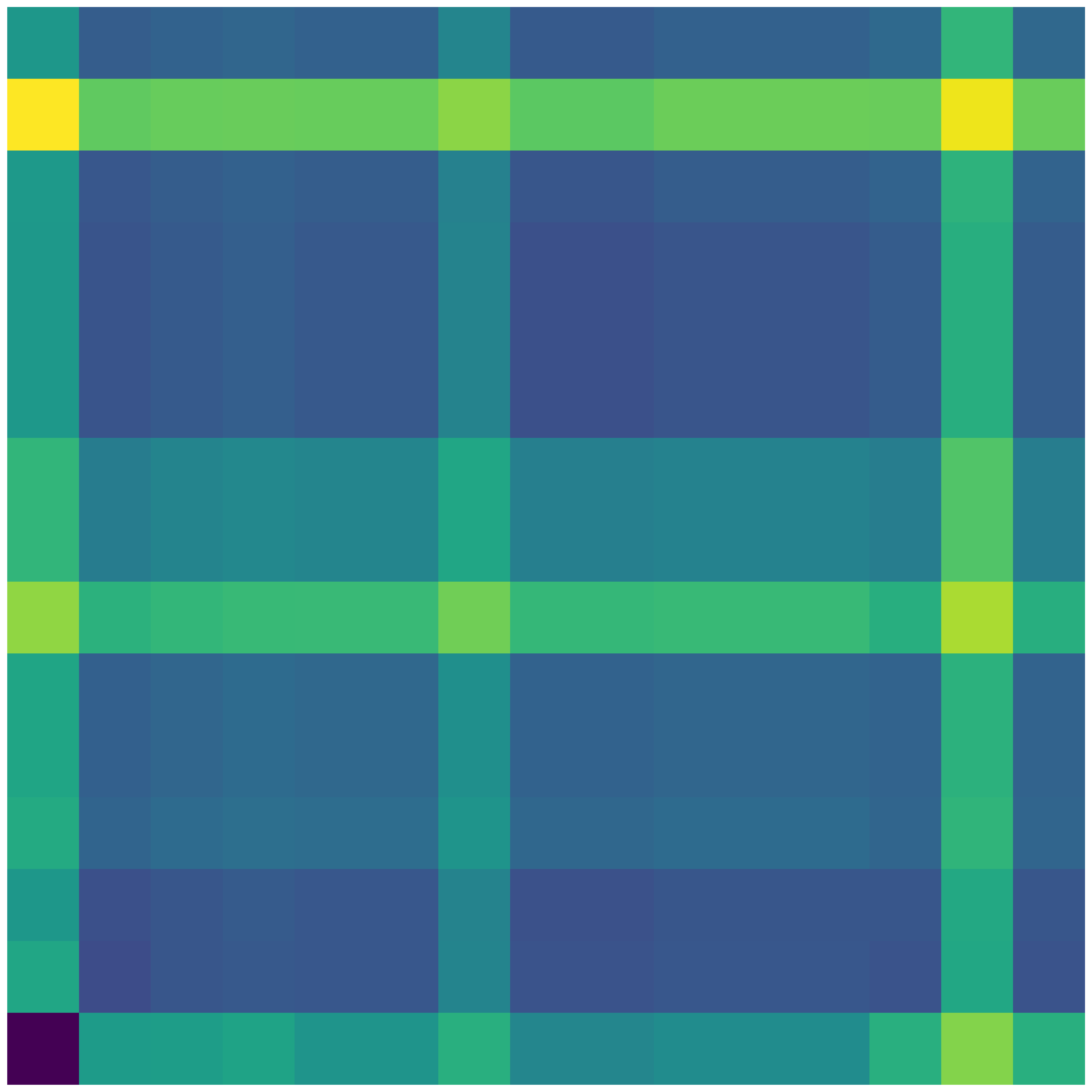}}
    \hfill
    \subfloat[VAE (4 cameras) vs. VAE (6 cameras)]{\includegraphics[height=2.8cm]{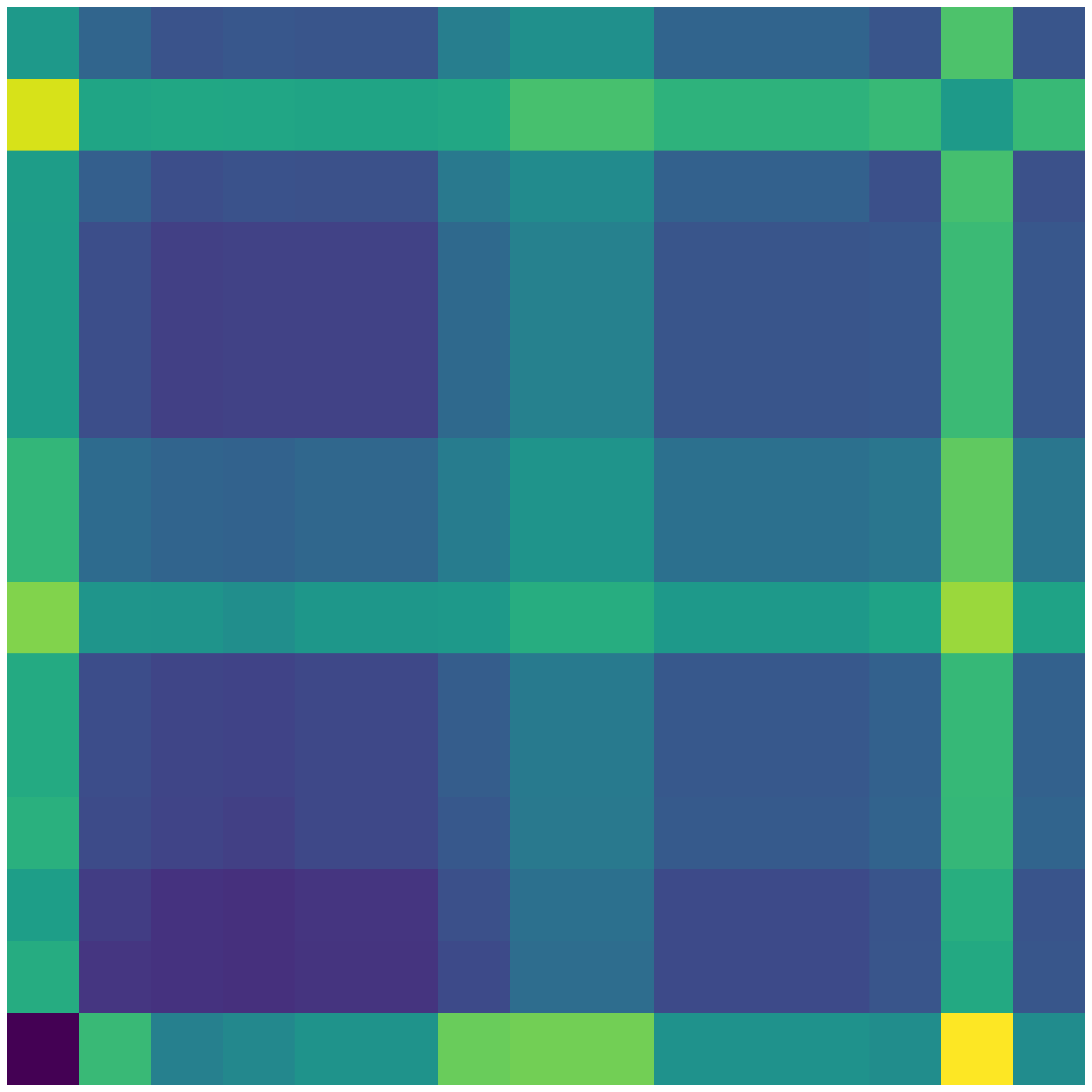}}
    \vfill
    \subfloat{\includegraphics[width=0.8\linewidth]{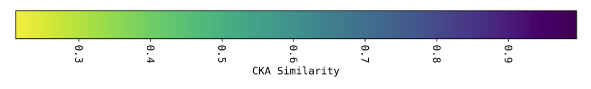}}
    \caption{
        CKA analysis of layer-wise representational similarity between perception models trained with the same supervisory signal.
        The heatmap compares latent features extracted from corresponding layers under the same input data.
        Deeper layers exhibit higher similarity, suggesting that the shared supervision encourages more consistent high-level representations.
    }\label{figure:cka-analysis}
\end{figure}

Empirical observations in Figure~\ref{figure:cka-analysis} further support this view.
Centered Kernel Alignment (CKA) measures the similarity between neural representations by comparing the geometry of their feature activations across the same set of input samples~\cite{kornblith2019similarity}.
In our analysis, we compare models trained with the same sensor inputs and the same supervisory signal.
Given an identical input batch, we extract the latent features produced at each layer and compute layer-wise CKA between the resulting activations.
The results show that representations learned under the same supervisory signal exhibit higher CKA similarity in later layers.
While high CKA similarity does not imply functional equivalence or exact feature-wise alignment, it suggests that the models develop consistent representational structures for organizing task-relevant factors.

\textbf{Assumption 3 (Supervisory Information Preservation).}
\textit{
Given a trained perception encoder $z = p_\phi(o)$, the latent representation approximately preserves the task-relevant information in $y$. Formally, there exists a decoder $q_\psi$ such that
\begin{equation}
    \mathbb{E}_{(o,y) \sim \mathcal{D}}
    \left[
    d_{\mathcal{Y}}([q_\psi\circ p_\phi](o), y)
    \right]
    \leq \epsilon,
\end{equation}
where $d_{\mathcal{Y}}$ is a task-dependent discrepancy measure, $\epsilon$ is small, and $\mathcal{D}$ denotes the task data distribution.
}

Together, Assumptions 1--3 and the observed representational consistency motivate the following hypothesis on task-local representation stitchability.

\textbf{Hypothesis (Policy-Compatible Representation Stitchability).}
\textit{
Consider two representations $z_1$ and $z_2$ learned under similar supervisory signals from paired observations of the same driving scenario instance.
If both representations retain task-relevant factors on the task distribution, then there can exist a low-complexity transformation $\mathcal{T}\colon \mathcal{Z}_2 \to \mathcal{Z}_1$ that maps $z_2$ into a latent space compatible with the downstream policy behavior.}

\textit{
In this work, low-complexity refers to transformations parameterized by either linear projections or shallow CNNs. Specifically, we empirically evaluate whether
\begin{equation}
    \mathbb{E}_{(o_1,o_2)\sim\mathcal{D}_{\mathrm{pair}}}
    \left[
    d_{\mathcal{A}}\left(
        [\pi_{\varphi_1}\circ \mathcal{T}\circ p_{\phi_2}](o_2),
        [\pi_{\varphi_1}\circ p_{\phi_1}](o_1)
    \right)
    \right]
    \leq \eta,
\end{equation}
holds approximately over the paired task distribution, where $\eta$ is a small behavior-mismatch error.
}

\textbf{Interpretation and scope.}
The above hypothesis should be interpreted as a task-level empirical hypothesis rather than a claim of global equivalence between representation spaces.
We do not assume that different latent spaces are globally isomorphic or information-complete.
Instead, the hypothesis only concerns whether representations trained under similar supervisory signals may admit practically useful alignments for downstream policy.
The hypothesis is therefore intentionally operational: its validity is evaluated by stitchability performance rather than by exact representational equivalence.


\subsection{Model Stitching}\label{section:model-stitching}

We use the following low-complexity model stitching methods to evaluate the stitchability hypothesis in Section~\ref{section:hypothesis}.

Given $m$ paired observations $\{(o_1^{(i)},o_2^{(i)})\}_{i=1}^m$ sampled from the same underlying environment states, two pretrained encoders $p_{\phi_1}$ and $p_{\phi_2}$ produce representations:
\begin{equation}
z_1^{(i)} = p_{\phi_1}(o_1^{(i)}), \quad
z_2^{(i)} = p_{\phi_2}(o_2^{(i)}),
\end{equation}
where $z_1^{(i)}, z_2^{(i)} \in \mathbb{R}^n$.
For convenience, we denote by $Z_1, Z_2 \in \mathbb{R}^{m \times n}$ the matrices obtained by stacking $\{z_1^{(i)}\}_{i=1}^m$ and $\{z_2^{(i)}\}_{i=1}^m$ row-wise.

To align representations across models, we introduce a stitching layer $\mathcal{T}(\cdot)$ that maps features from the representation space of $p_{\phi_2}$ to that of $p_{\phi_1}$.

\subsubsection{Linear Stitching}

We align vector representations using a linear transformation~\cite{lenc2015understanding}:
\begin{equation}
\mathcal{T}_{\mathrm{aff}}(z_2)=z_2A^\top+b,
\end{equation}
where $A\in\mathbb{R}^{n\times n}$ and $b\in\mathbb{R}^{1\times n}$ are estimated from the paired anchors~\cite{maiorca2023latent}.
The parameters are optimized via least squares:
\begin{equation}\label{equation:linear-stitching}
\min_{A,b}
\left\|
Z_2A^\top+\mathbf{1}_m b-Z_1
\right\|_F^2,
\end{equation}
where $\|\cdot\|_F$ denotes the Frobenius norm and
$\mathbf{1}_m\in\mathbb{R}^{m\times 1}$ is a column vector of ones.
This objective is equivalent to minimizing the sum of squared errors over all paired representations and admits a closed-form solution.

\subsubsection{Convolutional Stitching}

We use a lightweight convolutional stitching layer~\cite{csiszarik2021similarity}:
\begin{equation}
\mathcal{T}_{\mathrm{conv}}(M)
=
\mathrm{BN}(M * \Theta_{\mathrm{conv}}),
\end{equation}
where $\Theta_{\mathrm{conv}}$ denotes the learnable convolution kernel, $M$ denotes the feature map, and $\mathrm{BN}$ denotes batch normalization.
In practice, a $1\times1$ convolution for channel-wise alignment is enough to preserve spatial resolution.

During training, the source encoder $p_{\phi_2}$ and the target downstream decoder $q_{\psi_1}$ are frozen, and only the stitching layer is optimized:
\begin{equation}
    \mathcal{L}_{\mathrm{stitch}} 
    = 
    \frac{1}{m} 
    \sum_{i=1}^m 
    \mathcal{L}_{\mathrm{task}}
    \left(
        [q_{\psi_1}\circ \mathcal{T} \circ p_{\phi_2}](o_2^{(i)}),
        y^{(i)}
    \right),
\end{equation}
where $\mathcal{L}_{\mathrm{task}}$ is the task-specific loss function, e.g., mean squared error for regression tasks or cross-entropy for classification tasks, and $y^{(i)}$ is the corresponding target label for the $i$-th observation.

Optionally, a feature alignment objective can be added:
\begin{equation}
    \mathcal{L}_{\mathrm{align}} 
    = 
    \frac{1}{m} 
    \sum_{i=1}^m 
    \left\|
    [\mathcal{T} \circ p_{\phi_2}](o_2^{(i)}) 
    - 
    p_{\phi_1}(o_1^{(i)})
    \right\|_F^2,
\end{equation}
which encourages the stitched representations to be close to the target model's representations.

The overall loss function for training the convolutional stitching layer is a weighted sum of the stitching loss and the feature alignment loss:
\begin{equation}\label{equation:convolutional-stitching}
\mathcal{L}
=
\mathcal{L}_{\mathrm{stitch}}
+
\lambda \mathcal{L}_{\mathrm{align}},
\end{equation}
where $\lambda\geq0$ controls the contribution of the feature-level alignment term.

During stitching, all pretrained model parameters are frozen and only $\mathcal{T}$ is optimized. 
For linear stitching, $\mathcal{T}$ is estimated from paired representations using Eq.~\ref{equation:linear-stitching}. 
For convolutional stitching, $\mathcal{T}$ is trained using supervisory signals according to Eq.~\ref{equation:convolutional-stitching}. 
The resulting composed model $\pi_{\varphi_1}\circ\mathcal{T}\circ p_{\phi_2}$ is evaluated according to the objective defined by Eq.~\ref{equation:problem-formulation}.


\section{Experiments}\label{section:experiments}

\subsection{Experiment Setup}

\subsubsection{Supervisory Signal}

Based on prior findings in representation learning for autonomous driving~\cite{hu2022st,hu2023planning,jiang2023vad,liu2023deep}, we identify three types of policy-relevant information commonly encoded in supervisory signals $y$ that support downstream policy learning:

\begin{enumerate}
    \item \textbf{Geometric Structure}: road topology, lane connectivity, and locations of objects.

    \item \textbf{Semantic Information}: categories of scene entities (vehicles, pedestrians, cyclists).

    \item \textbf{Spatial Alignment}: correspondence between semantic entities and their physical locations.
\end{enumerate}

\begin{table}[htb]
\centering
\small
\setlength{\tabcolsep}{4pt}
\caption{Qualitative comparison of policy-relevant information preserved by common supervisory signals for downstream policy learning.}
\label{table:semantic-geometric-representation}

\begin{tabular}{lcccc}
\toprule
\textbf{Supervision} 
& \textbf{Geom.} 
& \textbf{Sem.} 
& \textbf{\makecell{Spatial\\Align.}} 
& \textbf{Rich.} \\
\midrule
BEV Segmentation & $\checkmark$ & $\checkmark$ & $\checkmark$ & Strong \\
3D Object Detection & $\sim$ & $\checkmark$ & $\checkmark$ & Moderate \\
Occupancy Grid & $\checkmark$ & $\times$ & $\checkmark$ & Moderate \\
Semantic Segmentation & $\sim$ & $\checkmark$ & $\sim$ & Moderate \\
Instance Segmentation & $\sim$ & $\checkmark$ & $\sim$ & Moderate \\
Depth Estimation & $\checkmark$ & $\times$ & $\sim$ & Moderate \\
Image Classification & $\times$ & $\sim$ & $\times$ & Weak \\
Natural Language & $\sim$ & $\sim$ & $\sim$ & Varies \\
\bottomrule
\end{tabular}

\vspace{1mm}
\footnotesize{
\textit{Geom.}: geometric scene structure; 
\textit{Sem.}: semantic information; 
\textit{Spatial Align.}: correspondence between semantic entities and physical locations. 
\textit{Rich.}: Information richness.
$\checkmark$: strong or explicit; 
$\sim$: partial or implicit; 
$\times$: little or none.
}
\end{table}

Table~\ref{table:semantic-geometric-representation} qualitatively compares common supervisory signals based on the extent to which they preserve these properties.
Among them, BEV segmentation and 3D object detection are selected because they represent two distinct supervision paradigms: dense scene-level and sparse object-centric representations. 
Evaluating model stitching under both settings enables validation across supervisory signals with different information richness and structural characteristics.

\subsubsection{Dataset Preparation}

\paragraph{Simulated data}
We construct two CARLA datasets with different sensor configurations and supervisory signals across varied traffic scenarios and weather conditions~\cite{dosovitskiy2017carla}.
Dataset statistics are reported in Table~\ref{table:dataset-setup}, and full sensor specifications are provided in Appendix~\ref{appendix:sensor-setup}.
To isolate factors that induce variation in perception representations and assess their impact on model stitching, the sensor setup of CARLA 2 is designed to approximate nuScenes. 
Unavoidable mismatches remain due to differences in data-collection platforms and simulator constraints.

\paragraph{Real-world data}
We use the nuScenes dataset~\cite{caesar2020nuscenes} with the official train/test splits for cross-domain evaluation.


\begin{table}[htb]
\centering
\caption{Dataset setup for model training and evaluation.}\label{table:dataset-setup}
\setlength{\tabcolsep}{4pt}
\begin{tabular}{l|ccccc}
\toprule[2pt]
\multirow{2}{*}{\textbf{Dataset}} & \multicolumn{2}{c}{\textbf{\# Samples}} & \multicolumn{2}{c}{\textbf{Sensor Setup}} & \multirow{2}{*}{\textbf{Supervision}} \\
\cmidrule(lr){2-5}
& Training & Testing & Camera & LiDAR & \\
\midrule
CARLA 1 & 39,768 & 10,752 & 4 & 1 & BEV Seg. \\
CARLA 2 & 39,768 & 10,752 & 6 & 1 & BEV Seg., 3D Det. \\
nuScenes & 28,130 & 6,019 & 6 & 1 & BEV Seg., 3D Det. \\
\bottomrule[2pt]
\end{tabular}

\vspace{1mm}
\footnotesize{
\textit{BEV Seg.}: BEV semantic segmentation;
\textit{3D Det.}: 3D object detection.
}
\end{table}

\subsubsection{Model Preparation}

This work builds upon RAMBLE~\cite{li2024imitation}, an end-to-end autonomous driving framework whose perception module is adapted from BEVFusion~\cite{liang2022bevfusion}. 
To investigate representation alignment under diverse perception updates, we train the perception module with controlled variations in sensor configuration, latent representation learning strategy, training dataset, and supervisory signal, as summarized in Table~\ref{table:model-setup}.

BEVFusion supports both BEV semantic segmentation and 3D object detection, enabling the study of representation learning under different supervision paradigms.
Specifically, BEV segmentation supervision is implemented using a VAE-based latent reconstruction architecture following the V-model design in RAMBLE, whereas 3D object detection supervision adopts a discriminative CNN-based latent prediction pipeline.
These settings represent dense scene-level and sparse object-centric supervision, respectively.

The downstream stack consists of a Transformer-based dynamics transition model and a Soft Actor-Critic (SAC) policy network.
Since the SAC policy operates on a compact latent state representation, downstream decision making is particularly sensitive to latent-space misalignment.
As a result, even moderate representation shifts can lead to substantial degradation in driving performance, making this setup a challenging testbed for evaluating compatibility restoration.


\begin{table}[htb]
\centering
\caption{Model setup for simulating different update conditions.}\label{table:model-setup}
\setlength{\tabcolsep}{4pt}
\begin{tabular}{l|cccc}
\toprule[2pt]
\textbf{Model ID} & \textbf{\makecell{Representation\\Model}} & \textbf{\makecell{Sensor\\Modality}} & \textbf{\makecell{Training\\Dataset}} & \textbf{Supervision} \\
\midrule
Model 1 & VAE & Image, LiDAR & CARLA 1 & BEV Seg. \\
Model 2 & VAE & Image, LiDAR & CARLA 1 & BEV Seg. \\
Model 3 & AE & Image, LiDAR & CARLA 1 & BEV Seg. \\
Model 4 & VAE & Image & CARLA 1 & BEV Seg. \\
Model 5 & VAE & LiDAR & CARLA 1 & BEV Seg. \\
Model 6 & VAE & Image, LiDAR & CARLA 2 & BEV Seg. \\
Model 7 & CNN + MLP & Image, LiDAR & CARLA 2 & 3D Det. \\
Model 8 & VAE & Image, LiDAR & nuScenes & BEV Seg. \\
Model 9 & CNN + MLP & Image, LiDAR & nuScenes & 3D Det. \\
\bottomrule[2pt]
\end{tabular}
\end{table}


\begin{table}[htb]
\centering
\small
\caption{Experimental case setup for evaluating model stitching under different update conditions.}\label{table:case-setup}
\setlength{\tabcolsep}{4pt}
\begin{tabular}{l|l|cc}
\toprule[2pt]
\textbf{Case} 
& \textbf{Updated Factor} 
& \textbf{Source} 
& \textbf{Target} \\
\midrule
Case 1 & Initialization & Model 2 & Model 1 \\
Case 2 & Rep. Model (VAE vs. AE) & Model 3 & Model 1 \\
Case 3 & Modality (Multi vs. Img.) & Model 4 & Model 1 \\
Case 4 & Modality (Multi vs. LiDAR) & Model 5 & Model 1 \\
Case 5 & Sensor Setup & Model 6 & Model 1 \\
Case 6 & Domain (Real $\to$ Sim., BEV) & Model 8 & Model 6 \\
Case 7 & Domain (Real $\to$ Sim., 3D Det.) & Model 9 & Model 7 \\
\bottomrule[2pt]
\end{tabular}
\end{table}

\subsubsection{Case Setup}

Table~\ref{table:case-setup} summarizes seven cases for evaluating model stitching under different update conditions.
\textbf{Source} denotes the model that provides the perception encoder for stitching, and \textbf{Target} denotes the model that provides the decoder and downstream policy model.
For interpretability, we organize the cases by expected representation shift.
These shift types are cumulative rather than mutually exclusive, such that higher-complexity settings include lower-level shifts.

\paragraph{Minimal shift (Cases 1--2).}
These sanity checks isolate low-discrepancy settings.
Case 1 measures sensitivity to random initialization under the same architecture.
Case 2 measures the effect of changing the latent objective (VAE to AE) with limited structural variation.

\paragraph{Modality/sensor shift (Cases 3--5).}
These cases vary perception inputs and sensor configuration.
Cases 3--4 compare single-modality and multi-modal perception.
Case 5 changes sensor count and pose, introducing larger geometric mismatch.

\paragraph{Domain shift (Cases 6--7).}
These cases use real-to-sim transfer: perception is trained on nuScenes, while downstream policy is evaluated in CARLA.
Case 6 evaluates closed-loop robustness under cross-domain representation mismatch.
Case 7 uses sparse 3D detection supervision and focuses on representation alignment rather than closed-loop driving metrics. 
Due to the high-dimensional convolutional feature maps in this setting, only convolutional stitching is applicable for alignment.
This setting examines whether task-agnostic supervisory signals can still induce transferable and stitchable latent representations across domains.

\subsubsection{Hardware Setup}

All experiments are conducted on a single server with one AMD Ryzen 9 7945HX CPU, \SI{64}{\giga\byte} RAM, and one NVIDIA RTX 4090 GPU.


\subsection{Experiment Results}

\subsubsection{Driving Performance}

Driving performance serves as the primary metric for validating our hypothesis.
Since the objective is to maintain the functionality of an end-to-end ADS after an upstream perception update, the quality of the resulting driving behavior directly reflects the effectiveness of the compatibility restoration method.
We compare three representative strategies to investigate whether lightweight transformations can recover downstream compatibility while achieving competitive driving performance relative to conventional update approaches.

\paragraph{Retraining}
After updating the perception model, all downstream components, including the dynamics transition model and the policy model, are retrained from scratch using the updated perception features.
All parameters are randomly initialized and optimized.
This setting serves as a strong reference baseline, representing the performance achievable when the entire downstream stack is re-optimized for the perception model, albeit at substantial computation costs.

\paragraph{Full fine-tuning}
After the perception model update, the downstream models are initialized from previously trained checkpoints.
All parameters remain trainable and are further optimized using the updated perception features.
Compared with retraining, this strategy leverages existing model weights to reduce training costs while still adapting the entire downstream stack.

\paragraph{Stitching}
We evaluate both linear and convolutional stitching methods for restoring compatibility between the updated perception model and the frozen downstream modules.
The linear stitcher is estimated using 2,048 sampled latent feature pairs.
The convolutional stitcher is implemented as a $1 \times 1$ convolution with 32 input and 32 output channels.
It is trained for a single epoch on data collected in CARLA using a batch size of 16 and a learning rate of 0.001.

Driving performance is evaluated in the high-fidelity CARLA simulator~\cite{dosovitskiy2017carla}, which provides realistic urban traffic scenarios and sensor simulations.
We select the EnterActorFlowV2 scenario from CARLA Leaderboard 2.0 as the evaluation task.
Following the data split protocol in~\cite{li2024imitation}, each route is approximately \SI{300}{m} long, consisting of 100--\SI{150}{m} before the scenario is triggered and 150--\SI{200}{m} afterward.
This scenario includes a non-signalized right turn followed by a merge into dense traffic from the left.
Successfully completing the route requires the vehicle to understand the surrounding traffic situation, anticipate the behavior of other road users, and make safe and efficient driving decisions.
Therefore, it provides a challenging and representative testbed for evaluating the impact of perception updates on downstream driving performance.

We adopt two standard evaluation metrics: Route Completion (RC) and Driving Score (DS).
RC measures the proportion of the predefined route successfully traversed by the vehicle, reflecting its ability to reach the destination.
DS is a composite metric that combines route completion with penalties for unsafe or improper behaviors, including collisions, traffic rule violations, and lane departures, thereby capturing both task success and driving quality.


\begin{table*}[htb]
\centering
\caption{Driving performance comparison under different restoration strategies.}\label{table:driving-performance}
\begin{tabular}{l|cc|cc|cc|cc|cc}
\toprule[2pt]
\multirow{2}{*}{\textbf{Case}} 
& \multicolumn{2}{c|}{\textbf{No Action}}
& \multicolumn{2}{c|}{\textbf{Retraining}}
& \multicolumn{2}{c|}{\textbf{Finetuning}}
& \multicolumn{2}{c|}{\textbf{Linear Stitching}}
& \multicolumn{2}{c}{\textbf{Convolutional Stitching}}
\\
\cmidrule{2-11}

& \textbf{RC} $\uparrow$
& \textbf{DS} $\uparrow$

& \textbf{RC} $\uparrow$
& \textbf{DS} $\uparrow$

& \textbf{RC} $\uparrow$
& \textbf{DS} $\uparrow$

& \textbf{RC} $\uparrow$
& \textbf{DS} $\uparrow$

& \textbf{RC} $\uparrow$
& \textbf{DS} $\uparrow$
\\
\midrule

No shift & 100.0 & 98.58 & - & - & - & - & - & - & - & - \\
Minimal shift (Case 1) & 53.87 & 34.91 & 100.0 & 99.12 & 100.0 & 96.24 & 98.21 & 96.00 & 100.0 & 98.51 \\
Sensor shift (Case 5) &  34.34 & 15.31 & 100.0 & 97.56 & 100.0 & 97.97 & 63.59 & 43.79 & 97.11 & 91.19 \\
Domain shift (Case 6) & 45.32 & 34.76 & 47.98 & 35.10 & 46.68 & 31.35 & 36.91 & 35.63 & 92.23 & 89.88 \\

\bottomrule[2pt]
\end{tabular}
\end{table*}


\begin{table*}[htb]
\centering
\caption{Update efficiency comparison across different restoration strategies (Case 5).}
\label{table:update-efficiency}
\begin{tabular}{l|c|cc|cc|cc|c}
\toprule[2pt]

\multirow{2}{*}{\textbf{Method}}
& \textbf{Updated Params}
& \multicolumn{2}{c|}{\textbf{Memory Usage}}
& \multicolumn{2}{c|}{\textbf{GPU Memory}}
& \multicolumn{2}{c|}{\textbf{Runtime}}
& \textbf{Interaction Step}
\\
\cmidrule{2-9}

& \textbf{(\%) $ \downarrow$}
& \textbf{(GB)$ \downarrow$}
& \textbf{(\%) $ \downarrow$}
& \textbf{(GB)$ \downarrow$}
& \textbf{(\%) $ \downarrow$}
& \textbf{(h)$ \downarrow$}
& \textbf{(\%) $ \downarrow$}
& \textbf{(\#) $ \downarrow$} \\\midrule

Retraining & 100.0 & 5.95 & 100.0 & 18.66 & 100.0 & 22.18 & 100.0 & 70,000 \\
Finetuning & 100.0 & 5.95 & 100.0 & 18.66 & 100.0 & 18.94 & 85.39 & 50,000 \\
\midrule
Linear stitching & \textbf{0.0} & \textbf{2.00} & \textbf{33.61} & \textbf{1.75} & \textbf{9.37} & \textbf{0.02} & \textbf{0.09} & \textbf{0} \\
Convolutional stitching & 0.001 & 2.35 & 39.50 & 5.66 & 30.33 & 0.91 & 4.10 & \textbf{0} \\

\bottomrule[2pt]
\end{tabular}
\end{table*}

Table~\ref{table:driving-performance} reveals several important observations about the impact of perception updates on driving performance.

When the sensor configuration remains unchanged, random initialization introduces only a mild feature-space shift (Case~1), a simple linear stitcher achieves an RC of 98.21 and a DS of 96.00, closely matching the performance of retraining and fine-tuning.
This result suggests that, under small distribution shifts, the original and updated latent representations remain approximately linearly aligned.

When the sensor configuration changes within the same simulator domain (Case~5), the representation discrepancy becomes substantially larger.
Under this setting, linear stitching recovers only part of the lost performance, reaching an RC of 63.59 and a DS of 43.79.
In contrast, convolutional stitching increases performance to an RC of 97.11 and a DS of 91.19, approaching the results achieved by retraining and fine-tuning, both of which attain nearly perfect RC and DS above 97.
These results indicate that sensor-induced representation shifts require a more expressive compatibility mapping than a simple linear transformation.

The most challenging scenario is the cross-domain setting (Case~6), where the perception model is trained on nuScenes while the downstream modules operate in CARLA. 
In this case, retraining and fine-tuning provide only marginal gains over the unadapted system, with RC remaining below 50 and DS remaining close to 35. 
Linear stitching similarly fails to improve performance, indicating that the representation discrepancy cannot be resolved through downstream adaptation or simple linear alignment. 
In contrast, convolutional stitching substantially improves performance, increasing RC from 45.32 to 92.23 and DS from 34.76 to 89.88.
Compared with the no-shift baseline, this corresponds to recovering more than 85\% of the performance lost due to domain shift.

Overall, these results indicate that model stitching is an effective mechanism for preserving driving performance across a wide range of perception updates.
While linear stitching is sufficient when the representation shift is small, larger sensor- and domain-induced shifts require a more expressive convolutional mapping.
Across all evaluated scenarios, convolutional stitching consistently restores performance to a level close to that achieved by retraining and fine-tuning, \textbf{suggesting that maintaining latent-space compatibility is important for preserving downstream driving behavior after perception updates.}

\subsubsection{Update Efficiency}

Beyond driving performance, our hypothesis also assumes that stitching enables efficient latent-interface repair with reduced computational and data requirements.
To evaluate adaptation efficiency, we consider the following metrics:

\paragraph{Updated parameters}
The proportion of original downstream model parameters updated during compatibility restoration.
Following our problem setting, the perception model is treated as an upstream perturbation and is therefore excluded from the parameter count.
For stitching-based methods, the original downstream modules remain frozen, and only the lightweight stitcher is estimated or optimized.

\paragraph{Memory usage}
Peak host memory consumption during adaptation, including model parameters, gradients, optimizer states, intermediate training buffers, and simulator overhead.

\paragraph{GPU memory usage}
Peak GPU memory consumption during adaptation, including model parameters, gradients, optimizer states, activation tensors for backpropagation, and simulator overhead.

\paragraph{Runtime}
Total wall-clock time required for compatibility restoration. Since simulator initialization and environment loading are non-negligible in end-to-end autonomous driving systems, the reported runtime includes both optimization time and simulator interaction overhead.

\paragraph{Interaction step}
The number of additional online environment interaction steps required to recover 90\% of the original driving performance.
Offline calibration samples used for estimating the stitcher are not counted as additional interaction steps.

Table~\ref{table:update-efficiency} summarizes the adaptation cost of different restoration strategies for Case 5.
Due to variations in sensor configurations and simulator workloads, the reported CPU and GPU memory usage in other cases may fluctuate by approximately \SI{2}{GB} across different runs.
Nevertheless, the relative differences between methods remain consistent.

\begin{figure*}
    \centering
    \subfloat[Ground truth]{\includegraphics[width=0.19\linewidth]{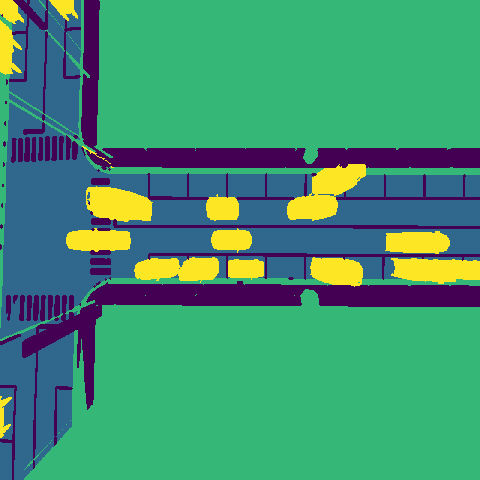}}
    \hfill
    \subfloat[Source (native decoder)]{\includegraphics[width=0.19\linewidth]{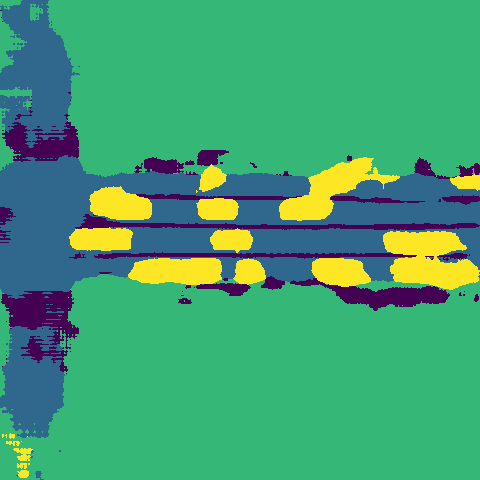}}
    \hfill
    \subfloat[Target (native decoder)]{\includegraphics[width=0.19\linewidth]{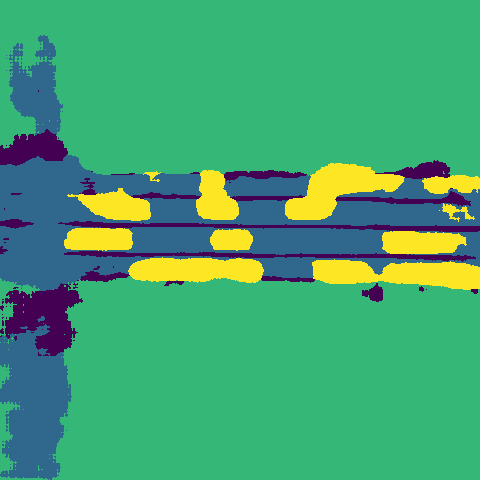}}
    \hfill
    \subfloat[Direct connecting (source $\to$ target)]{\includegraphics[width=0.19\linewidth]{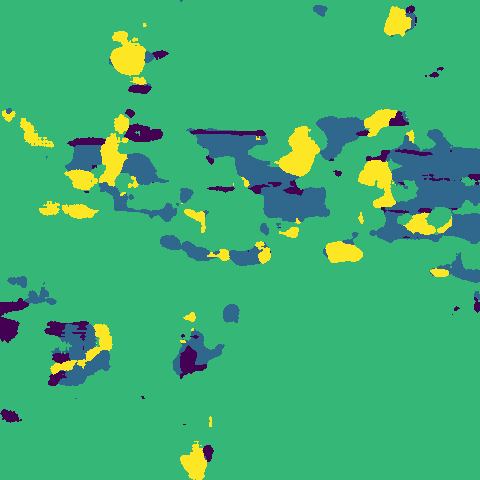}}
    \hfill
    \subfloat[Linear stitching (source $\to$ target)]{\includegraphics[width=0.19\linewidth]{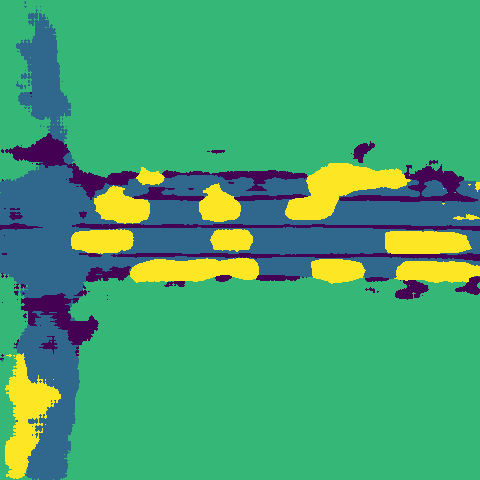}}
    \caption{Visualization of Case 2 using the linear stitcher for AE-to-VAE stitching under BEV segmentation supervision on CARLA 1.}\label{figure:segmentation-stitch-visualization}
\end{figure*}

\begin{figure*}
    \centering
    \subfloat[Ground truth]{\includegraphics[width=0.19\linewidth]{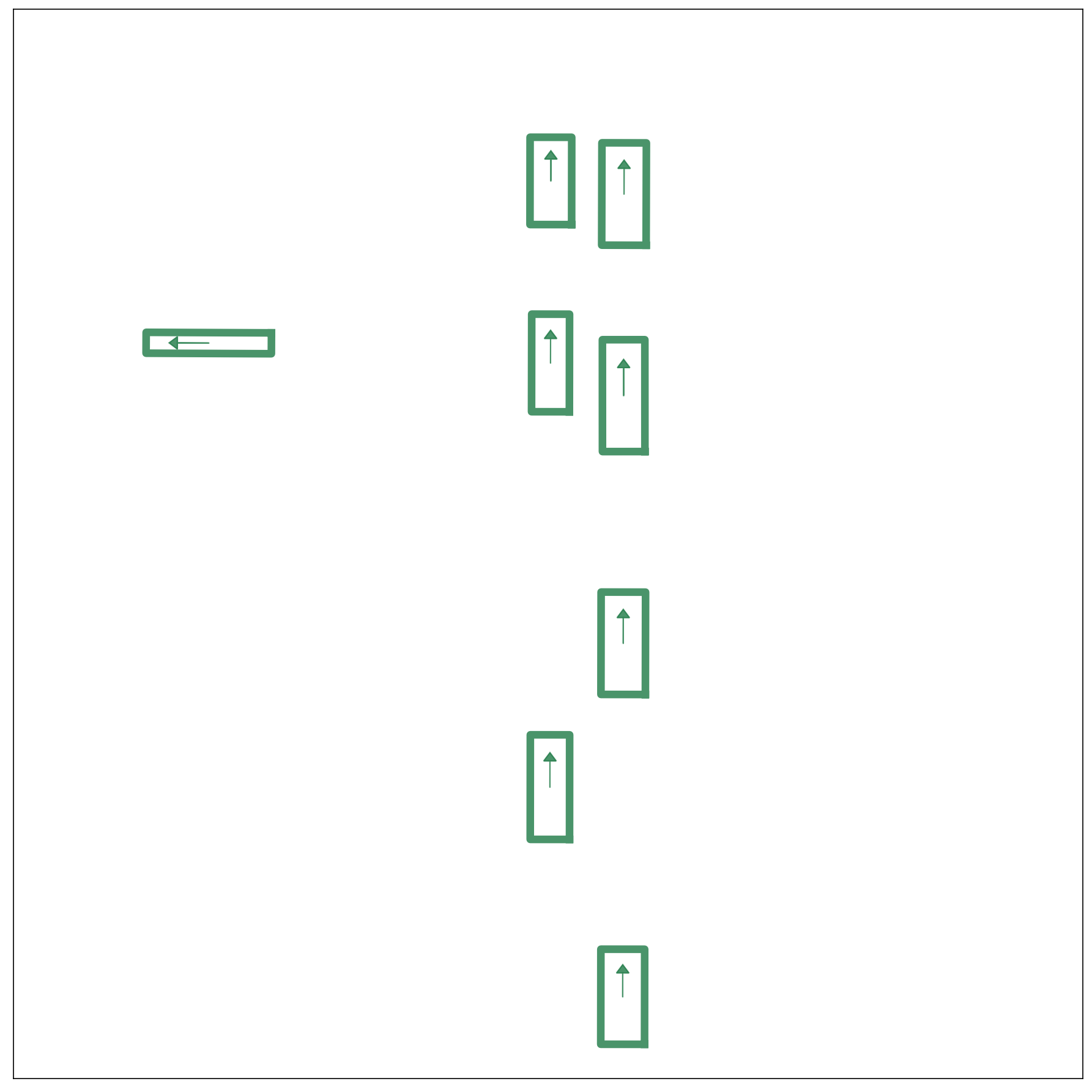}}
    \hfill
    \subfloat[Source (native decoder)]{\includegraphics[width=0.19\linewidth]{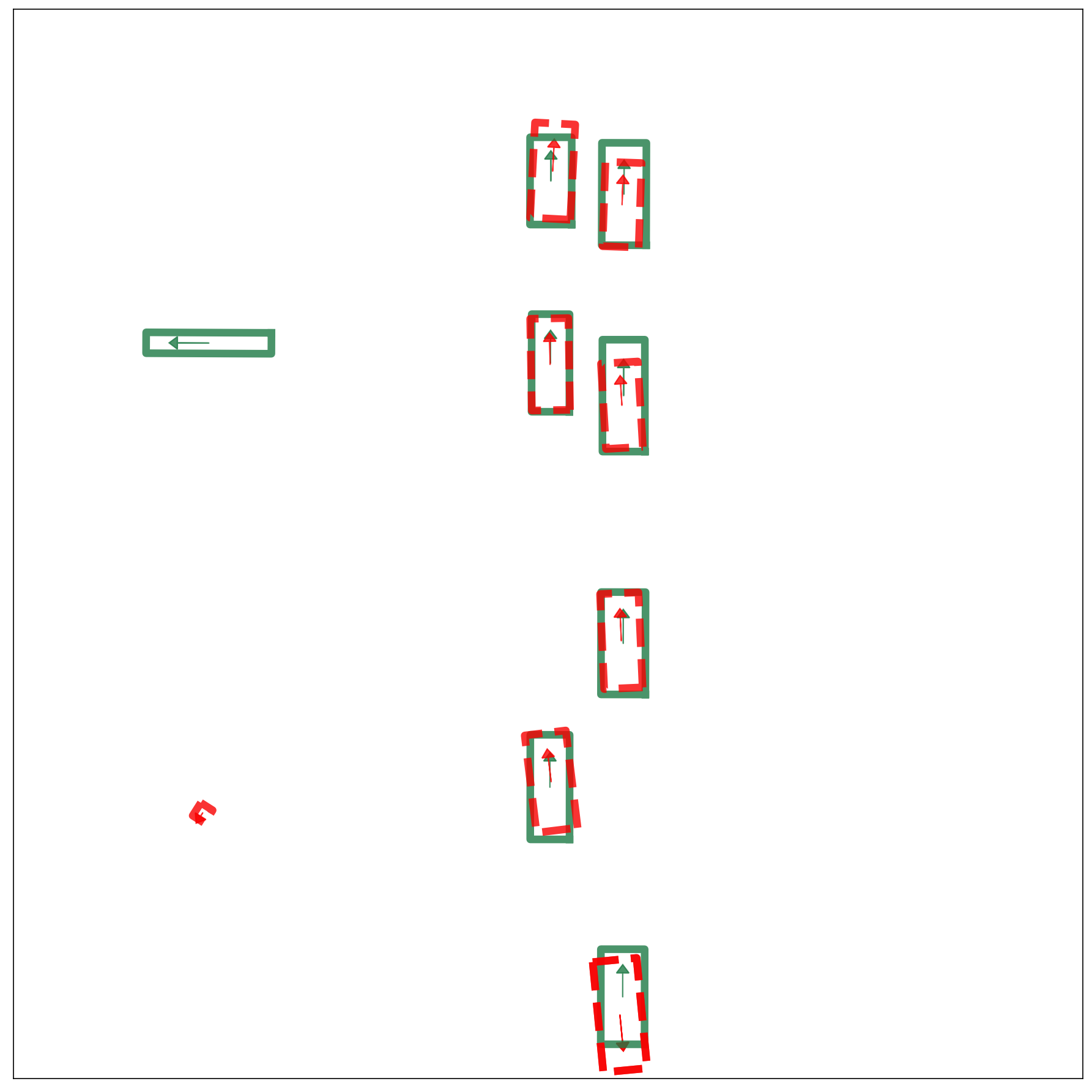}}
    \hfill
    \subfloat[Target (native decoder)]{\includegraphics[width=0.19\linewidth]{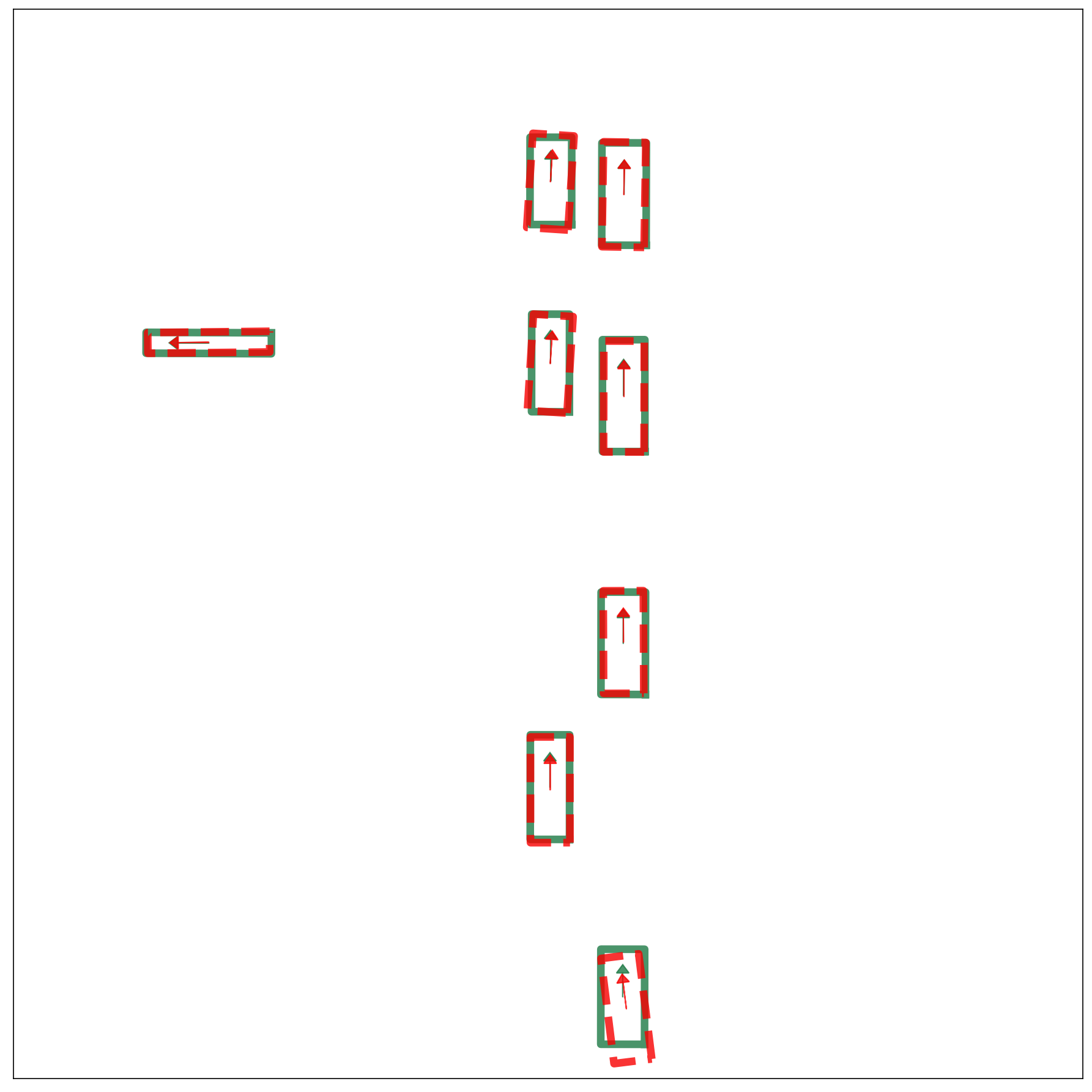}}
    \hfill
    \subfloat[Direct connecting (source $\to$ target)]{\includegraphics[width=0.19\linewidth]{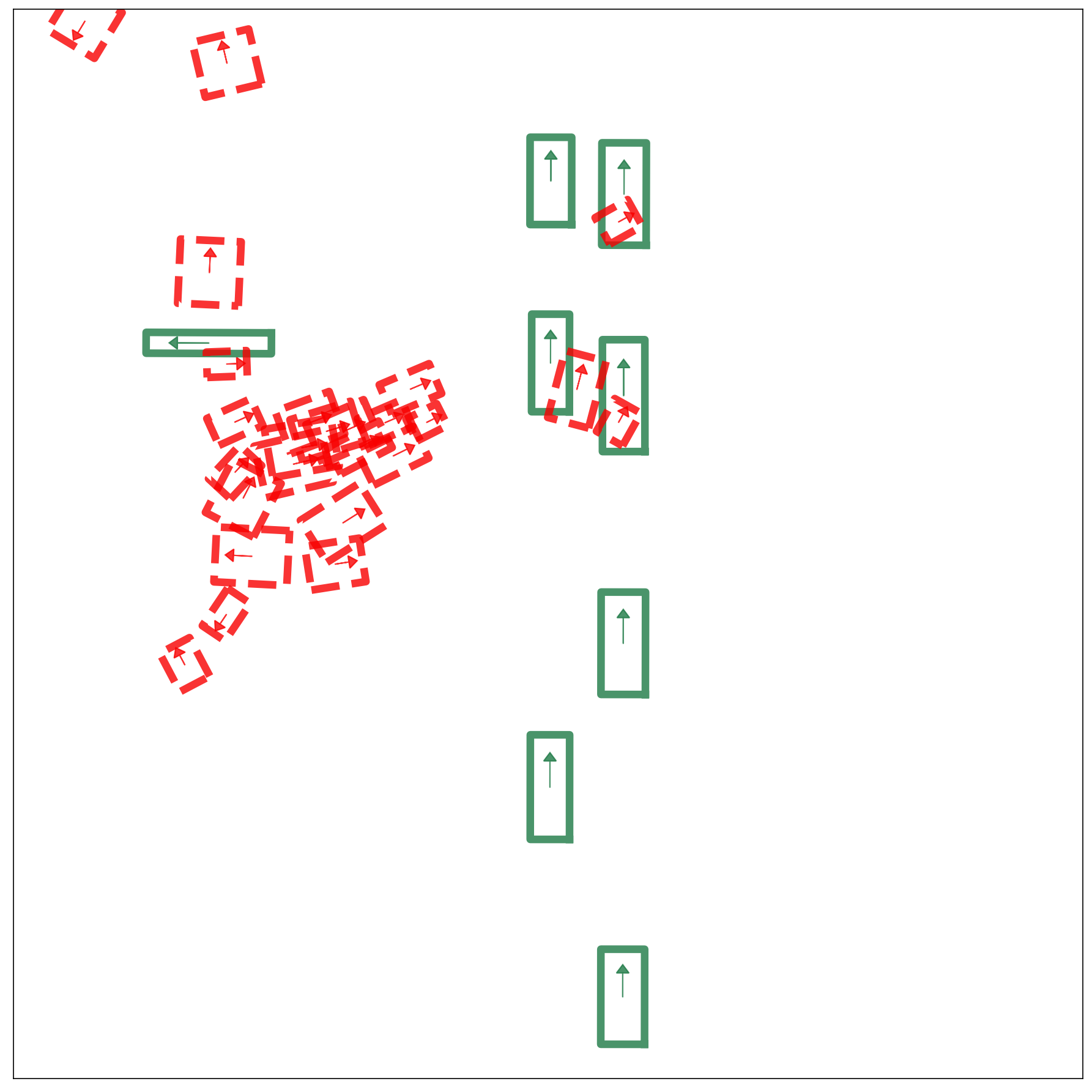}}
    \hfill
    \subfloat[Convolutional stitching (source $\to$ target)]{\includegraphics[width=0.19\linewidth]{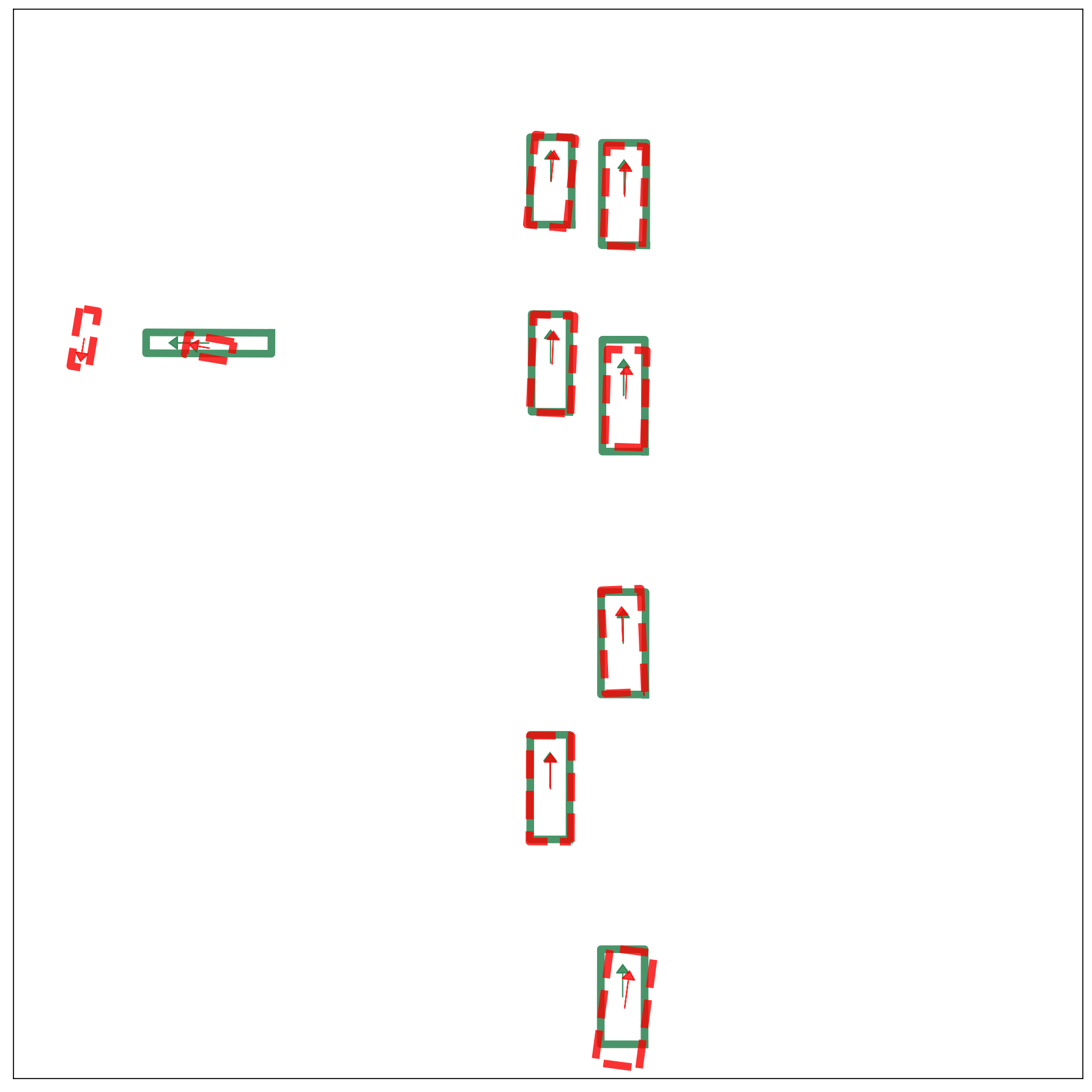}}
    \caption{Visualization of Case 7 using the convolutional stitcher for VAE-to-VAE stitching under BEV object detection supervision, with different sensor positions and rotations between the source and target models. The source model was trained on nuScenes, and the target model was trained on CARLA 2.}\label{figure:detection-stitch-visualization}
\end{figure*}

The most notable advantage of stitching lies in its adaptation efficiency.
Compared with retraining, linear stitching reduces the adaptation time from \SI{22.18}{h} to \SI{0.02}{h}, corresponding to a speedup of more than three orders of magnitude.
Convolutional stitching requires only \SI{0.91}{h}, achieving a 24$\times$~reduction in runtime.
Similarly, GPU memory consumption decreases substantially from \SI{18.66}{GB} for retraining to \SI{1.75}{GB} and \SI{5.66}{GB} for linear and convolutional stitching, respectively.
These reductions are achieved because stitching avoids backpropagation through the downstream driving models and only estimates a lightweight interface transformation.

The parameter efficiency gains are even more pronounced.
Retraining and fine-tuning require updating the entire downstream stack, whereas convolutional stitching updates only 0.001\% of the downstream parameters and linear stitching does not modify any original downstream parameters.
As a result, stitching provides a lightweight mechanism for restoring compatibility without re-optimizing the downstream models.

Another important observation is the elimination of additional online interaction.
Retraining and fine-tuning require 70,000 and 50,000 environment interaction steps, respectively, to recover driving performance.
In contrast, both stitching methods restore compatibility without requiring additional online exploration, relying solely on offline latent-space alignment.
This significantly reduces the cost of updating a deployed ADS, where data collection and simulator interaction often dominate the adaptation pipeline.

Despite these substantial reductions in computational and data requirements, the driving performance achieved by both stitching methods remains comparable to that of retraining and fine-tuning.
This result suggests that recovering compatibility does not necessarily require re-learning the downstream driving policy.
Instead, a lightweight transformation at the latent interface is often sufficient to restore effective communication between the updated perception module and the frozen downstream models.

\subsubsection{Representation Alignment Analysis}

To further evaluate representation alignment, we measure the quality of stitched intermediate representations using the downstream perception metrics of the target decoder.
We use mIoU for semantic segmentation and mAP for object detection.
The qualitative examples in Figures~\ref{figure:segmentation-stitch-visualization} and~\ref{figure:detection-stitch-visualization} further show that stitching helps recover semantic structure in segmentation and preserve localized cues in detection.

Table~\ref{table:alignment-analysis} reports the quantitative results. 
Cases 1--5 evaluate stitching within dataset CARLA 1, while Cases 6--7 evaluate cross-domain transfer from nuScenes to dataset CARLA 2.

\begin{table*}[htb]
\centering
\caption{Perceptual evaluation of representation alignment under different stitching methods.}\label{table:alignment-analysis}
\begin{tabular}{l|c|cc|ccc}
\toprule[2pt]
\textbf{Case} 
& \makecell{\textbf{Metric}}
& \makecell{\textbf{Source} \\ \textbf{(native decoder)}}
& \makecell{\textbf{Target} \\ \textbf{(native encoder)}}
& \makecell{\textbf{Direct Connecting} \\ \textbf{(source $\to$ target)}} 
& \makecell{\textbf{Linear Stitching} \\ \textbf{(source $\to$ target)}}
& \makecell{\textbf{Convolutional Stitching} \\ \textbf{(source $\to$ target)}}
\\
\midrule
Case 1 & mIoU $\uparrow$ & 0.586 & 0.593 & 0.203 & 0.567 & 0.563 \\
Case 2 & mIoU $\uparrow$ & 0.604 & 0.593 & 0.207 & 0.575 & 0.566 \\
Case 3 & mIoU $\uparrow$ & 0.556 & 0.593 & 0.188 & 0.515 & 0.522 \\
Case 4 & mIoU $\uparrow$ & 0.575 & 0.593 & 0.191 & 0.526 & 0.514 \\
Case 5 & mIoU $\uparrow$ & 0.601 & 0.593 & 0.186 & 0.313 & 0.542 \\
Case 6 & mIoU $\uparrow$ & 0.178 & 0.601 & 0.183 & 0.390 & 0.499 \\
Case 7 & mAP@50 $\uparrow$ & 0.011 & 0.623 & 0.000 & \textemdash & 0.391 \\
\bottomrule[2pt]
\end{tabular}

\vspace{1mm}
\footnotesize{\textemdash: Linear stitching not applicable due to high-dimensional convolutional feature maps.}
\end{table*}

In all cases, directly connecting the source encoder to the target decoder leads to substantial performance degradation.
When the shift in perception representations is mild (Cases 1--4), linear stitching can effectively recover most of the performance, indicating that the representation shift is largely linear and can be corrected by a simple affine transformation.
In Case 5, which involves more camera views, linear stitching is less effective, suggesting stronger representation misalignment.
Still, convolutional stitching recovers most of the performance.

A similar trend is observed in the cross-domain settings Case 6 and 7.
The source models (Model 8 and Model 9) achieve reasonable performance in their original domain (mIoU of 0.501 and mAP@50 of 0.466), but suffer significant degradation when evaluated on CARLA 2 due to distribution shift.
Directly transferring the source encoder results in features that are hardly interpretable.
Nevertheless, both linear and convolutional stitching substantially improve performance without modifying the trained models, indicating that transferable semantic information remains accessible after correcting feature-space misalignment.
This observation is consistent with prior findings on model stitching~\cite{kornblith2019similarity}.


\section{Conclusion}

This paper investigated how to effectively preserve downstream driving behavior after perception updates in end-to-end ADS. 
Motivated by the observation that perception models are frequently updated while downstream decision-making modules are expensive to retrain and revalidate, we introduced the Policy-Compatible Representation Stitchability Hypothesis and evaluated it through a series of controlled perception updates involving changes in sensor configuration, representation model, and training dataset.

Across the seven evaluated settings, the results provide consistent empirical support for the hypothesis that policy-compatible latent representations often admit low-complexity alignments.
In particular, lightweight stitching modules were able to restore downstream driving performance without modifying models.
These findings hold for a downstream stack combining Transformer-based world modeling and reinforcement learning-based control, indicating that the proposed compatibility restoration mechanism may be applicable beyond a specific policy implementation.
The results are consistent with the interpretation that \textbf{performance degradation following perception updates is largely attributable to latent-space misalignment, and that task-relevant information may be largely recoverable through appropriate transformations.}
Even in the most challenging cross-domain setting, convolutional stitching improved driving score from 34.76 to 89.88, recovering most of the performance lost after the perception update.

Beyond effectiveness, stitching offers substantial practical advantages.
Compared with retraining-based adaptation, convolutional stitching reduces adaptation time from \SI{22.18}{h} to \SI{0.91}{h} while updating only 0.001\% of downstream parameters.
Moreover, it restores compatibility without requiring any additional online interaction, eliminating the 70,000 environment steps required by retraining.
These reductions are achieved while maintaining driving performance close to that of retraining and fine-tuning.
Such properties make stitching particularly attractive for deployed ADS, where retraining costs and safety validation requirements can be substantial.

In summary, our findings suggest that latent-space compatibility should be treated as a first-class design consideration when evolving perception models in end-to-end ADS.
Rather than retraining the entire driving stack after every perception update, restoring compatibility through lightweight latent-space alignment provides an effective and scalable alternative. We hope this work encourages further investigation of representation compatibility as a foundation for modular and continuously evolving ADS.

\bibliographystyle{IEEEtran}
\bibliography{main}

\begin{IEEEbiography}[{\includegraphics[width=1in,height=1.25in,clip,keepaspectratio]{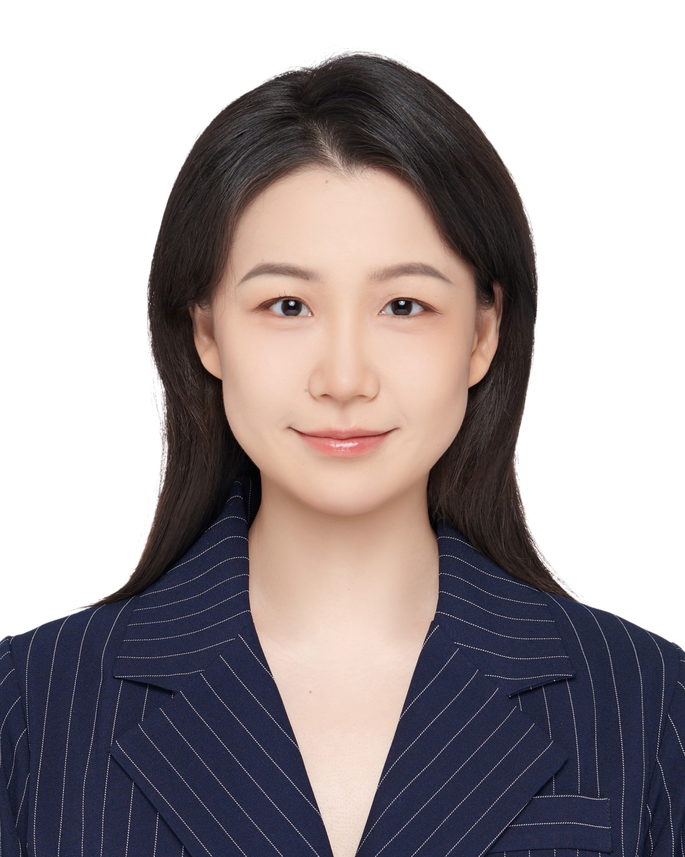}}]{Yueyuan LI}
    received a Bachelor's degree in Electrical and Computer Engineering from the University of Michigan-Shanghai Jiao Tong University Joint Institute in 2020. She is pursuing a Ph.D. degree in Automation from Shanghai Jiao Tong University. Her main fields of interest are the security of the autonomous driving system and driving decision-making. Her research activities include reinforcement learning, behavior modeling, and simulation.
\vspace{-20pt}
\end{IEEEbiography}

\begin{IEEEbiography}[{\includegraphics[width=1in,height=1.25in,clip,keepaspectratio]{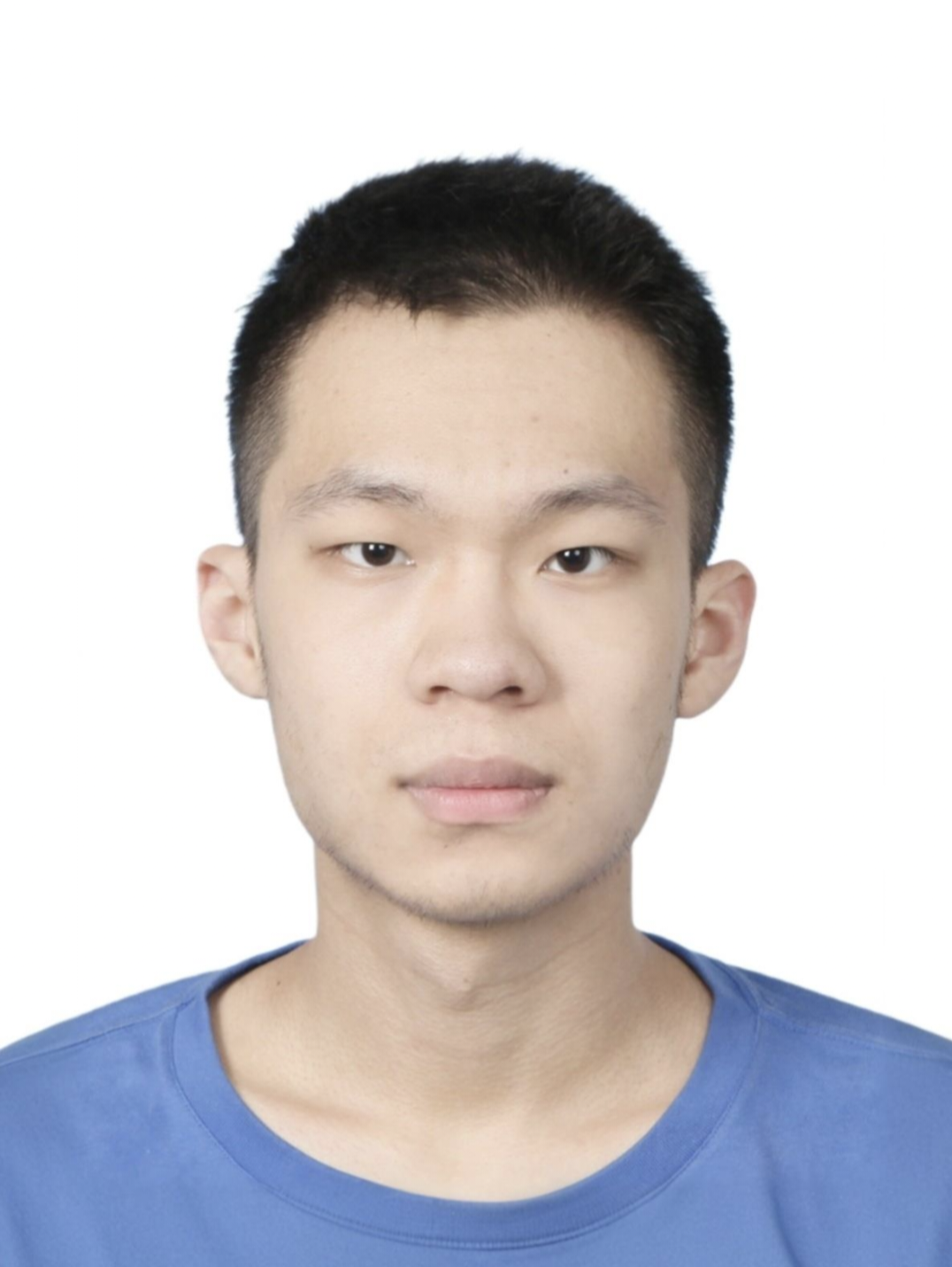}}]{Yifei XIAO}
     received a Bachelor's degree in Automation from Shanghai Jiao Tong University in 2026. He is currently pursuing the Master's degree in Automation from Shanghai Jiao Tong University. His main research interests include representation learning and reinforcement learning.
\vspace{-20pt}
\end{IEEEbiography}

\begin{IEEEbiography}
[{\includegraphics[width=1in,height=1.25in,clip,keepaspectratio]{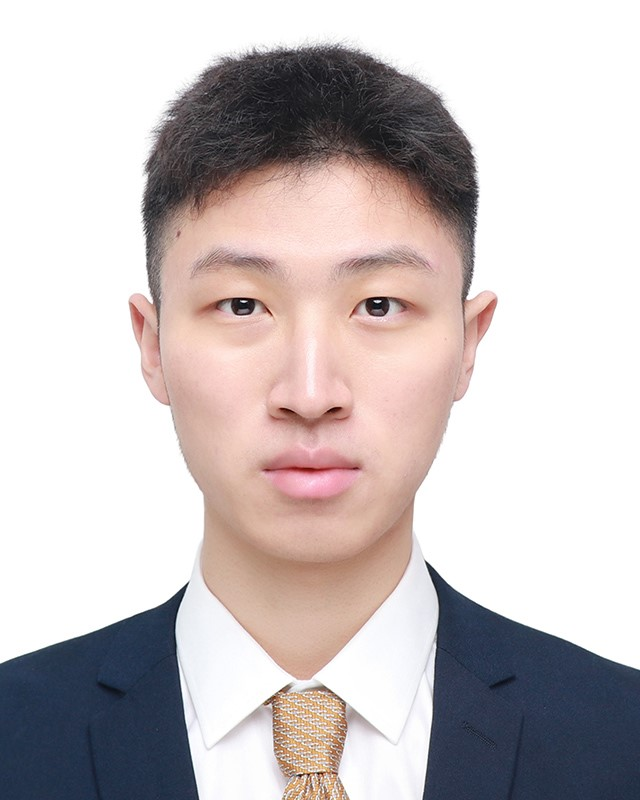}}]{Mingyang JIANG} 
    received a Bachelor's degree in engineering from Shanghai Jiao Tong University in 2023, and a Master's degree in Control Science and Engineering from Shanghai Jiao Tong University in 2026. His main research interests are end-to-end planning, driving decision-making, and reinforcement learning for autonomous vehicles.
\vspace{-20pt}
\end{IEEEbiography}

\begin{IEEEbiography}
[{\includegraphics[width=1in,height=1.25in,clip,keepaspectratio]{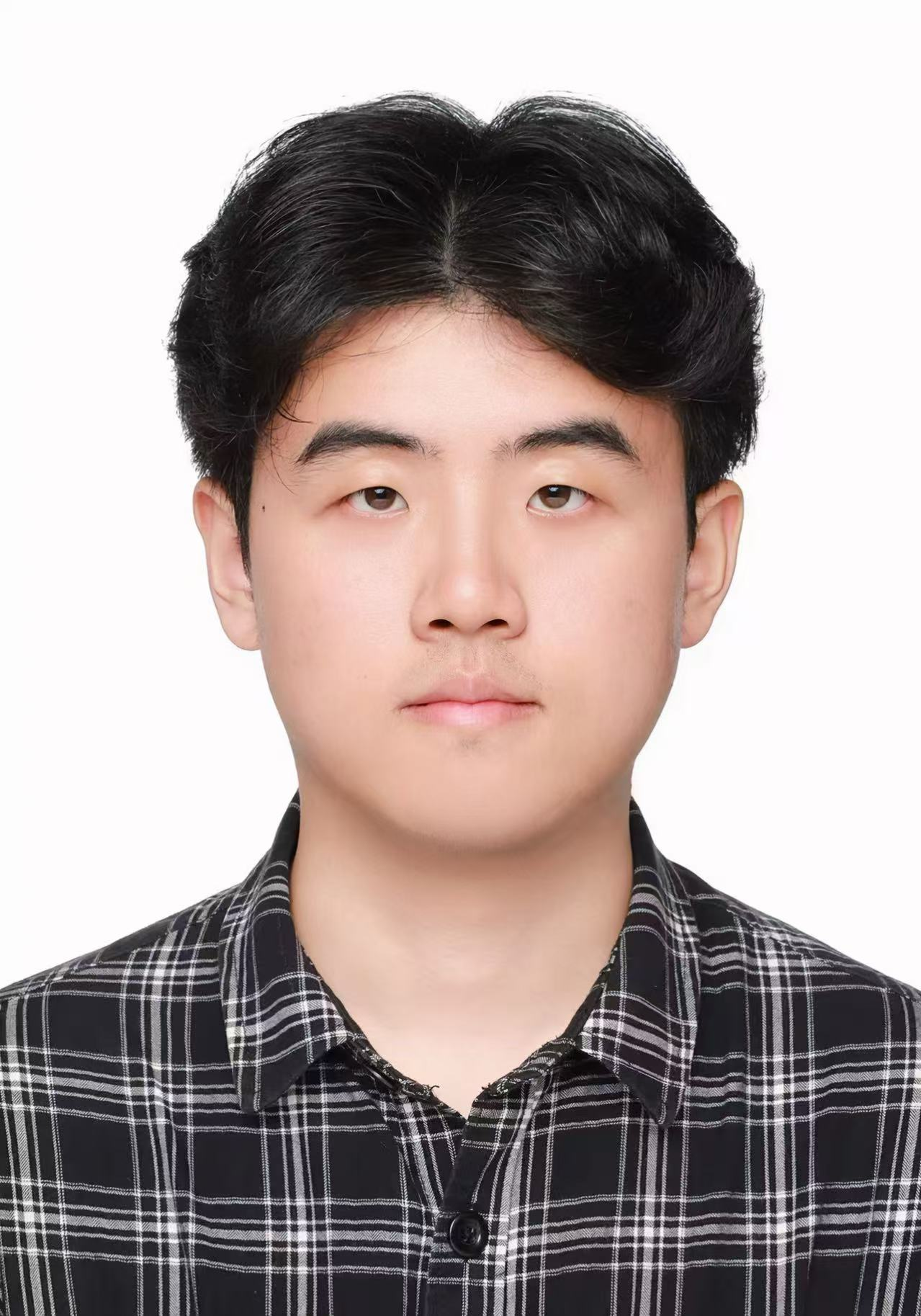}}]{Xiang ZUO} 
    received a Bachelor's degree in mathematics from Shanghai Jiao Tong University in 2025. He is currently pursuing the Ph.D. degree in mechanical engineering with Shanghai Jiao Tong University. His research interests include artificial intelligence for scientific discovery and autonomous systems.
\vspace{-20pt}
\end{IEEEbiography}

\begin{IEEEbiography}[{\includegraphics[width=1in,height=1.25in,clip,keepaspectratio]{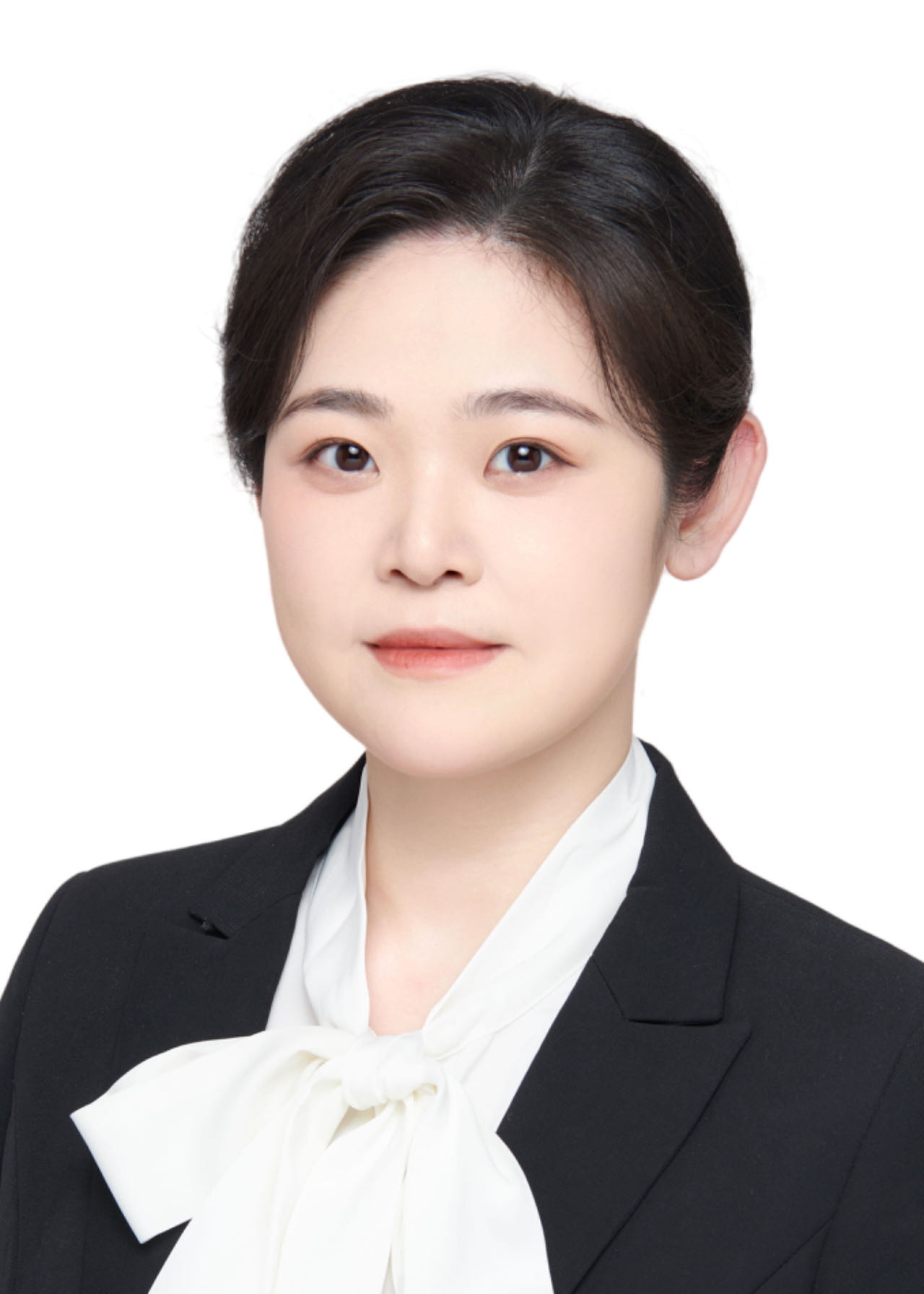}}]{Songan ZHANG}
    received B.S. and M.S. degrees in automotive engineering from Tsinghua University in 2013 and 2016, respectively. Then, she went to the University of Michigan, Ann Arbor, and received a Ph.D. in mechanical engineering in 2021. After graduation, she worked as a research scientist on the Robotics Research Team at Ford Motor Company. Presently, she is an assistant professor at the Global Institute of Future Technology (GIFT) at Shanghai Jiao Tong University. Her research interests include accelerated evaluation of autonomous vehicles, model-based reinforcement learning, and meta-reinforcement learning for autonomous vehicle decision-making.
\vspace{-20pt}
\end{IEEEbiography}



\begin{IEEEbiography}[{\includegraphics[width=1in,height=1.25in,clip,keepaspectratio]{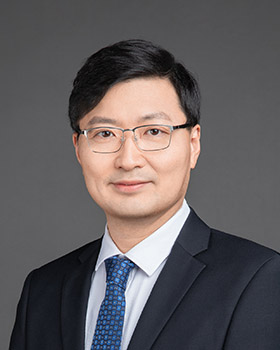}}]{Ming YANG}
    received his Master’s and Ph.D. degrees from Tsinghua University, Beijing, China, in 1999 and 2003, respectively. Presently, he holds the position of Distinguished Professor at Shanghai Jiao Tong University, also serving as the Director of the Innovation Center of Intelligent Connected Vehicles. Dr. Yang has been engaged in the research of intelligent vehicles for more than 25 years.
\vspace{-20pt}
\end{IEEEbiography}

\vspace{30pt}

\newpage
\appendix


\section{Sensor Setup}\label{appendix:sensor-setup}

The experimental sensor suite consists of multiple RGB cameras and a LiDAR sensor.
Models 1–5 use the configuration reported in Table~\ref{table:sensor-configuration-1} with an image resolution of 640$\times$480 pixels, whereas Models 6–7 use the configuration in Table~\ref{table:sensor-configuration-2} with an image resolution of 800$\times$450 pixels.
Models 8–9 further adopt the camera intrinsic parameters from the nuScenes dataset while preserving the same camera orientations and FOV settings as in Table~\ref{table:sensor-configuration-2}.

Note that CARLA defines sensor locations relative to the center of the ego vehicle, whereas nuScenes defines sensor locations relative to the rear-axle center.
Therefore, the sensor translation parameters (\textit{x}, \textit{y}, \textit{z}) are not strictly aligned across the two platforms, although the camera orientations and viewing angles remain equivalent.


\begin{table}[htb]
    \centering
    \caption{The configured sensor's location with respect to the center of the ego vehicle.}
    \label{table:sensor-configuration-1}
    \begin{tabular}{l|c c c c c c c}
    \toprule[2pt]
        \textbf{Sensor} & x (m) & y (m) & z (m) & roll & pitch & yaw & FOV \\ \midrule
        Camera (front) & 0.2 & 0.0 & 1.8 & 0$^{\circ}$ & 0$^{\circ}$ & 0$^{\circ}$ & 120$^{\circ}$ \\
        Camera (left) & -0.1 & -0.4 & 1.8 & 0$^{\circ}$ & -15$^{\circ}$ & -90$^{\circ}$ & 120$^{\circ}$ \\
        Camera (right) & -0.1 & 0.4 & 1.8 & 0$^{\circ}$ & -15$^{\circ}$ & 90$^{\circ}$ & 120$^{\circ}$ \\
        Camera (back) & -0.5 & 0.0 & 1.8 & 0$^{\circ}$ & 0$^{\circ}$ & 180$^{\circ}$ & 120$^{\circ}$ \\
        LiDAR & 0.0 & 0.0 & 1.8 & 0$^{\circ}$ & 0$^{\circ}$ & 0$^{\circ}$ & 360$^{\circ}$ \\
        \bottomrule[2pt]
    \end{tabular}
\end{table}

\begin{table}[htb]
    \centering
    \caption{The configured sensor's location with respect to the center of the ego vehicle.}
    \label{table:sensor-configuration-2}
    \begin{tabular}{l|c c c c c c c}
    \toprule[2pt]
        \textbf{Sensor} & x (m) & y (m) & z (m) & roll & pitch & yaw & FOV \\ \midrule
        Camera (front) & 1.45 & 0.0 & 1.60 & 0$^{\circ}$ & 0$^{\circ}$ & 0$^{\circ}$ & 70$^{\circ}$ \\
        \makecell[l]{Camera\\(front left)} & 1.20 & -0.55 & 1.60 & 0$^{\circ}$ & 0$^{\circ}$ & -55$^{\circ}$ & 70$^{\circ}$ \\
        \makecell[l]{Camera\\(front right)} & 1.20 & 0.55 & 1.60 & 0$^{\circ}$ & 0$^{\circ}$ & 55$^{\circ}$ & 70$^{\circ}$ \\
        \makecell[l]{Camera\\(back left)} & -1.10 & 0.55 & 1.60 & 0$^{\circ}$ & 0$^{\circ}$ & -110$^{\circ}$ & 70$^{\circ}$ \\
        \makecell[l]{Camera\\(back right)} & -1.10 & 0.55 & 1.60 & 0$^{\circ}$ & 0$^{\circ}$ & 110$^{\circ}$ & 70$^{\circ}$ \\
        Camera (back) & -1.35 & 0.0 & 1.60 & 0$^{\circ}$ & 0$^{\circ}$ & 180$^{\circ}$ &110$^{\circ}$ \\
        LiDAR & 0.0 & 0.0 & 1.80 & 0$^{\circ}$ & 0$^{\circ}$ & 0$^{\circ}$ & 360$^{\circ}$ \\
        \bottomrule[2pt]
    \end{tabular}
\end{table}

\end{document}